\pdfoutput=1

\documentclass[10pt,journal,compsoc]{IEEEtransaction/IEEEtran}
\usepackage{graphicx}
\usepackage{url}
\usepackage{amsmath,amssymb} % define this before the line numbering.

\newcommand{\fig}[1]{Figure~\ref{fig:#1}}
\newcommand{\sect}[1]{Section~\ref{sect:#1}}
\newcommand{\tab}[1]{Table~\ref{tab:#1}}

\newcommand{\eq}[1]{(\ref{eq:#1})}

\ifCLASSOPTIONcompsoc
  \usepackage[nocompress]{cite}
\else
  \usepackage{cite}
\fi
\ifCLASSINFOpdf
\else
\fi
\begin{document}
\title{Instance Segmentation by Deep Coloring}

\author{Victor~Kulikov,
        Victor~Yurchenko,
        and Victor~Lempitsky% <-this % stops a space
\IEEEcompsocitemizethanks{
\IEEEcompsocthanksitem V. Kulikov is with the Center for Computational and Data-Intensive Science and Engineering (CDISE),  Skolkovo Institute of Science and Technology (Skoltech), Skolkovo, Moscow, Russia\protect\\
E-mail: v.kulikov@skoltech.ru
\IEEEcompsocthanksitem V. Yurchenko is with Yandex, Moscow, Russia \protect\\
\IEEEcompsocthanksitem V. Lempitsky is with Samsung AI Center, Moscow and with Skolkovo Institute of Science and Technology (Skoltech)\protect\\
} % <-this % stops an unwanted space
\thanks{Manuscript received July 23, 2018}}


\markboth{July 2018}%
{Kulikov \MakeLowercase{\textit{et al.}}: Instance Segmentation by Deep Coloring}



\IEEEtitleabstractindextext{%
\begin{abstract}
We propose a new and, arguably, a very simple reduction of instance segmentation to semantic segmentation. This reduction allows to train feed-forward non-recurrent deep instance segmentation systems in an end-to-end fashion using architectures that have been proposed for semantic segmentation. Our approach proceeds by introducing a fixed number of labels (colors) and then dynamically assigning object instances to those labels during training (coloring). A standard semantic segmentation objective is then used to train a network that can color previously unseen images. At test time, individual object instances can be recovered from the output of the trained convolutional network using simple connected component analysis. In the experimental validation, the coloring approach is shown to be capable of solving diverse instance segmentation tasks arising in autonomous driving (the Cityscapes benchmark), plant phenotyping (the CVPPP leaf segmentation challenge), and high-throughput microscopy image analysis.

The source code is publicly available: \url{https://github.com/kulikovv/DeepColoring}. 
\end{abstract}

\begin{IEEEkeywords}
Instance segmentation, Semantic segmentation, Graph coloring, Convolutional neural networks
\end{IEEEkeywords}}

\maketitle

\IEEEdisplaynontitleabstractindextext

% \fi
\IEEEpeerreviewmaketitle

\IEEEraisesectionheading{\section{Introduction}\label{sec:introduction}}
%\section{Introduction}

\if@twocolumn%
  \begin{figure}[t]
    \begin{center}
    \includegraphics[width=\textwidth,keepaspectratio]{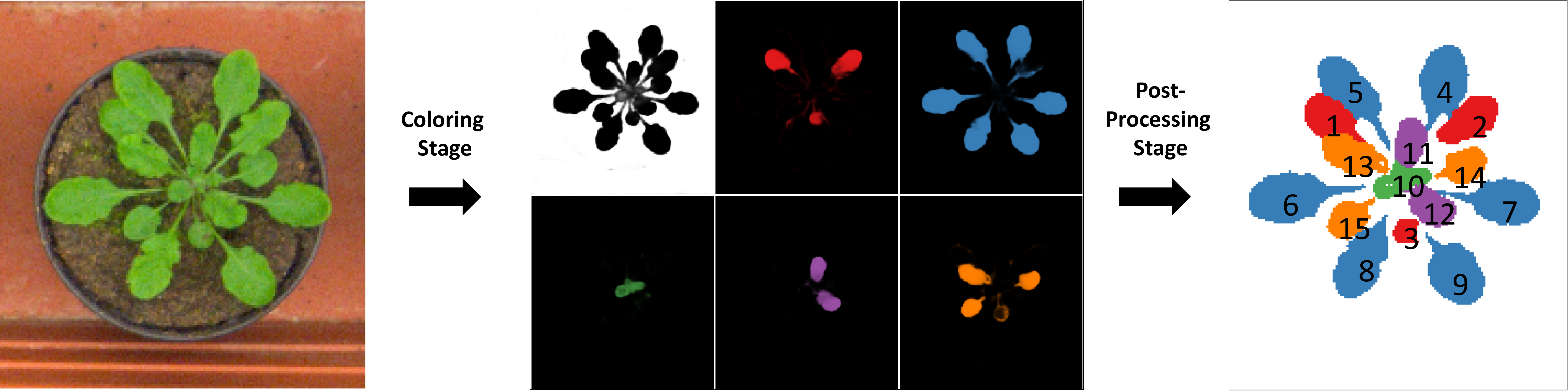}
    \end{center}
    \caption{ \textbf{Instance segmentation by Deep Coloring} at test time. The input image is passed through a coloring network with fixed number (six in this example) of output channels. As the last layer of the network corresponds to pixel-wise softmax, every pixel effectively ends up colored in one of six colors. During training, the coloring process ensures that the network tends to use different colors for close-by instances. As a result, at test time the instances can be recovered by assigning each pixel to the maximal color label and then using connected component analysis on the predicted label maps. Overall, in this example 15 instances (plus background) are recovered, while only six color labels are used by the system.}
    \label{fig:teaser}
    \end{figure}
\else% \@twocolumnfalse
  \begin{figure*}[t]
    \begin{center}
    \includegraphics[width=\textwidth,keepaspectratio]{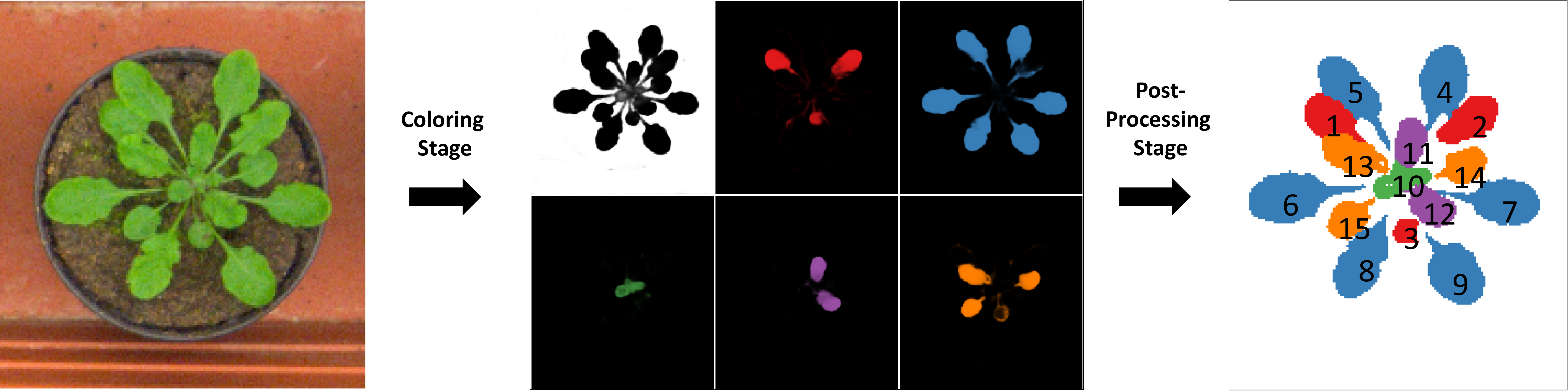}
    \end{center}
    \caption{ \textbf{Instance segmentation by Deep Coloring} at test time. The input image is passed through a coloring network with fixed number (six in this example) of output channels. As the last layer of the network corresponds to pixel-wise softmax, every pixel effectively ends up colored in one of six colors. During training, the coloring process ensures that the network tends to use different colors for close-by instances. As a result, at test time the instances can be recovered by assigning each pixel to the maximal color label and then using connected component analysis on the predicted label maps. Overall, in this example 15 instances (plus background) are recovered, while only six color labels are used by the system.}
    \label{fig:teaser}
    \end{figure*}
\fi

%\textit{Instance segmentation}
\IEEEPARstart{I}{nstance segmentation} is the problem of identifying and outlying individual instances of one or several semantic classes in an image. The number of instances are usually not known in advance, and the visual appearance of the instances may be very similar. Instance segmentation is therefore often regarded as a harder counterpart to the \textit{semantic segmentation} problem that attributes pixels to a known number of classes that usually have distinct visual appearance. The latter problem can now be solved rather successfully using end-to-end trained deep learning architectures that usually ``mix and match'' ideas of fully convolutional networks \cite{long2015fully}, the ``hourglass'' U-net architecture \cite{ronneberger2015u}, dilated convolutions \cite{yu2015multi}, separable convolutions \cite{rigamonti2013learning,chollet2016xception} and several others in order to achieve maximum speed and/or accuracy. In general, semantic segmentation using feed-forward convolutional networks trained in an end-to-end fashion is working well \cite{cordts2016cityscapes,lin2014microsoft}. 

At the same time, there are few feed-forward instance segmentation architectures that can be trained in an end-to-end manner \cite{romera2016recurrent,he2017mask}, and such architectures are also considerably more complex than semantic segmentation architectures. The extra complexity leads to longer training times, longer inference times as well as to the difficulties in re-implementation and parameters tuning. The extra complexity of instance segmentation over semantic segmentation can often be attributed to the fact that there are no consistent ordering between object instances in each training or test images that can be utilized by convolutional architectures in a natural way.

In this work, we propose a simple way to reduce instance segmentation to semantic segmentation that allows to train instance segmentation in an (almost) end-to-end fashion. In a nutshell, our idea is to use a segmentation network that assigns pixels to a \textit{constant} number of labels (that we call colors) in order to segment a \textit{variable} number of instances. Essentially, the task of the deep network in our approach is to \textit{color} multiple instances using fixed set of colors. Thus, the same color label can be reused for multiple instances  in the same image as long as such instances are spatially separated (\fig{teaser}). Thus, reusing single label for multiple instances allows our approach to handle arbitrary and potentially large number of class instances in an image.

Unlike semantic segmentation, where the assignment of pixels to classes is given, we do not assign each instances to specific colors in advance, as any such hard assignment is likely to be suboptimal. Instead, the coloring process is dynamic, in the sense that the coloring of instances is performed on-the-fly every time a training image is revisited by a stochastic learning process and the segmentation loss for this image is computed. Consequently, same instance in the same image may change its color between subsequent training epochs. 

During training, we use a simple coloring rule that effectively assigns each instance a color that leads to a small semantic segmentation loss, while also enforcing that pixels of adjacent instances are not colored with the same color. Thus, the loss computation in our training procedure is reminiscent of the graph coloring process (with instances loosely corresponding to graph vertices), which justifies the name of our approach. As the coloring process forces adjacent instances to be colored differently, recovering outlines of individual instances from the resulting coloring at test time can be accomplished using simple component analysis with few parameters. This connected component analysis is the only part of our approach that is not included in the otherwise end-to-end training process.

Importantly, our reduction can utilize standard semantic segmentation architectures (such as U-Net\cite{ronneberger2015u}) and many of the ideas from semantic segmentation can be directly transfered to our approach. The versatility of the approach is demonstrated through the validation on three rather diverse datasets corresponding to an autonomous driving, a plant phenotyping and a microscopy image analysis applications.

%two different datasets, namely the plant phenotyping dataset \cite{scharr2014annotated,minervini2016finely}, which is a standard benchmark for instance segmentation consisting of overhead photos of plant leaves and the Cityscapes benchmark for autonomous driving \cite{cordts2016cityscapes}. In both cases, we demonstrate the ability of the proposed reduction to discern individual instances and to achieve competitive results.

% Описать задачу
% Описание задачи: желательно привести пример
% Instance segmentation is the problem of detecting and delineating each different object of interest appearing in the image. 
		
% We present a simple and flexible end-to-end framework for object instance segmentation of biomedical data. The model is based on semantic segmentation framework U-net \cite{ronneberger2015u}, trained using a custom loss function in order to split neighboring objects into different feature maps. 

% Идеи статьи:
% 1. Потановка задачи в виде семантической сегментации на разные карты end-to-end
% 2. Функция потерь с границей
% 3. Методика обучения с понижением стохастического шума в картах
%The developed instance segmentation framework should:
%be able to handle images with a large number of objects
%get a reliable quality for a relative small number of training examples 
%have a high speed in training and inference phases
\section{Related work}\label{sec:related}
%\section{Related Work}

\textbf{Semantic segmentation} has seen considerable progress in recent years. Most of this progress was associated with new architectures of underlying deep convolutional networks. One of the overarching ideas is finding new ways to enlarge receptive fields of output neurons in the way that is efficient both in the computational and statistical sense. Having large receptive field is also very important for the success of the coloring network within our method. The popular ways of increasing the receptive field is using deeper neural network \cite{he2016deep}, adding dilated/``a trous'' convolution \cite{yu2015multi} and using "hourglass" architectures \cite{ronneberger2015u,paszke2016enet} that include downscaling/upscaling parts with skip connections. In \cite{zhao2016pyramid} the  features from different scales are concatenated to produce better segmentation. Our method can work on the top of any semantic segmentation architecture and can benefit of all the tricks used in those methods.         

\textbf{Proposal-based instance segmentation methods} are based on pixelwise refinement of object proposals. Some of the recent methods decompose the instance segmentation into detection and binary segmentation parts \cite{dai2016instance,li2016fully,he2017mask}. Those methods can be trained end-to-end with a non-maximum suppression during the post-processing. A similar approach described in  \cite{cordts2016cityscapes,hariharan2014simultaneous,chen2015multi,cordts2016cityscapes} where the method \cite{arbelaez2014multiscale} is used to generate category independent region proposals, followed by a classification step.  Proposal-based methods show the best performance in instance segmentation competitions like MS COCO~\cite{lin2014microsoft}, Pascal VOC2012~\cite{Everingham10} and CityScapes~\cite{cordts2016cityscapes}. They are at the same time limited by the quality of the object detection routine, which is hard to train on small datasets and for objects not approximated well by bounding boxes. Our method does not require on any object proposals or bounding box detection, although it can be integrated with such method.

\textbf{Recurrent instance segmentation methods} use recurrent neural networks to generate the instances sequentially one-by-one. Romera \textit{et al.}~\cite{romera2016recurrent} train a network for end-to-end instance segmentation and counting using LSTM \cite{hochreiter1997long}. The network produces a segmentation mask and a confidence value for each instance. Based on confidence the masks are kept or rejected from the final solution. In~\cite{ren2016end} a combination of recurrent networks with bounding box proposals is used. Their framework is composed of four major parts: external memory, that keeps the current state of the segmentation, LSTM based bounding box prediction network, a separated semantic segmentation network and a score network to estimate the confidence of the detected instance. This method gives the current state-of-the-art on the leaf segmentation task on the plant phenotyping dataset. Our method outperforms \cite{romera2016recurrent} but lags behind \cite{ren2016end} on the CVPPP benchmark. At the same time, our approach enjoys the fact that run-time is almost independent on the number of objects in the image.

\textbf{Proposal-free methods} usually operate by breaking the output of semantic segmentation into instances (semantic segmentation may be precomputed or performed in a parallel stream to instance segmentation). Usually, an extra information is predicted at each pixel that assists in the breaking. Thus, deep watershed transform~\textit{et al.}~\cite{bai2016deep} learns to predict directions and the energy of watershed transform. In \cite{uhrig2016pixel} a template matching scheme for instance segmentation were proposed. Their architecture includes three parts that perform semantic segmentation, depth estimation, and angle estimation for each pixel. Template matching algorithm predicts instances using the output of these three branches.

De Brabandere~\textit{et al.}~\cite{de2017semantic} is arguably most similar to ours. It uses metric learning to obtain high-dimensional pixel embeddings that map pixels of the same instance close to each other, while mapping pixels from different instances further apart. To retrieve instances from the embeddings, a clustering algorithm is applied. Our approach is similar in spirit to \cite{de2017semantic}, but simplifies the pipeline, as it replaces metric learning and clustering-based postprocessing, with classification learning and connected component-based postprocessing. In the experimental section, we perform several comparisons with \cite{de2017semantic}.

%Kirillov \textit{et al.} \cite{kirillov2016instancecut} combine semantic segmentation with a predicted edge map to produce instances using multi-cut procedure.
%Our method falls into the proposal free category and it also predicts auxiliary information at each pixel by assigning it to one of the color classes. This can be used to break apart the precomputed semantic segmentation mask. Notably, it can also avoid running semantic segmentation as a separate process (e.g.\ we show that coloring and figure-ground segmentation can be performed in one process on CVPPP dataset).

\textbf{Weakly-supervised semantic segmentation} approaches such as \cite{Xu15,Zhang15,Kolesnikov16} learn by alternating the updates of the target pixel labels (based on the current prediction of the segmentation labels and the constraints imposed by week labeling) and the updates of the network parameters. Such dynamic modification of the target pixel labels is also performed within our approach, which is however solving a different problem (instance segmentation).

\section{Instance Segmentation via Coloring}\label{sec:method}
\newcommand{\X}{{\mathbf{x}}}
\newcommand{\Y}{{\mathbf{y}}}
\newcommand{\Z}{{\mathbf{z}}}
\newcommand{\M}{{\mathbf{M}}}
\newcommand{\B}{{\mathbf{b}}}
\newcommand{\Mn}{\M_\text{halo}}
\renewcommand{\wp}{w_{+}}
\newcommand{\wn}{w_{-}}

We now discuss our approach in detail. In this work, we consider instance segmentation that does not assign instances to semantic classes.  In this case, the task is simply to assign every pixel in an image to either one of the object instances or to background. The generalization for \textit{semantic instance segmentation} that combines instance separation with semantic segmentation is relatively straightforward, at least for small number of classes.

\subsection{Inference}

At test time, processing a previously unseen color image $ \X$ of size $w \times h \times 3$ is performed in two stages  (\fig{teaser}) . Firstly in the \textbf{coloring stage}, the image is mapped by a coloring network $\Psi$ with learned parameters $\theta$ into a new image $\Y = \Psi(\X;\theta)$ with the same spatial size and $C$ channels (i.e.\ $\Y \in \mathbb{R}^{C \times W \times H}$). We denote with $\Y[c]$ the $c$-th channel of the map $\Y$ and with $\Y[c,p]$ the value at the spatial position $p$ in this channel.

In our experiments, unless noted otherwise, $\Psi$ has a U-net-style architecture \cite{ronneberger2015u} with an ``hourglass'' encoder-decoder shape augmented with skip connections across the bottleneck. The exact architectures are detailed in the experimental section. The U-net architecture is characterized by very large receptive fields of the output neurons, which is important for instance segmentation. 

We assume that the final layer of the network is the softmax operation, so that the outputs of the network for any image are non-negative and sum to one at every spatial location (i.e.\ $\Y > \mathbf{0}$ and $\sum_c \Y[c,p] = 1$ for any $p$), and can thus be interpreted as the probability of a pixel to take a certain color. We reserve the first color to the background, i.e.\ interpret pixels $p$ with high $\Y[1,p]$ as background.

The \textbf{post-processing stage} in our approach assigns each pixel to the color of the highest probability resulting in a map $\Z$ of the size $W\times H$ with $\Z[p] \in \{1,\dots,C\}$. We then find all connected components in $\Z$, and treat the non-background connected components that are bigger than a certain size threshold $\tau$ as object instances. The choice of the size threshold is discussed in \sect{params}. The connected components that are smaller than the size threshold are reassigned to background. 

Finally, due to the nature of the learning process (discussed below), when two connected components of the same color are encountered in close proximity, this provides an evidence that they correspond to the same object instance. We therefore optionally merge instances assigned to the same color, if the Hausdorf distance between the respective connected components is less than the proximity threshold $\rho$. Such merging allows our method to recover objects that are disconnected in the image because of partial occlusion.

%treat the first output channel $\Y[0]$ as the background probability channel and the remaining channels $\Y[i], i>0$ as object channels. We perform thresholding of each of these $D$ channels and consider the connected components of the thresholded images which are big enough as object instances. More precisely, for each channel $\Y[i], i>0$ we consider the binary map $\B[i] = \Y[i] > \tau_i$ and we identify all connected components of $\B[i]$ that are greater than $\sigma_i$ pixels as object instances. The parameters $\tau_i$ and $\sigma_i$ characterize the $i$-th channel and are set using validation on the validation set.

\subsection{Learning}

% \begin{figure}
%     \begin{center}
%     \includegraphics[width=0.8\linewidth,keepaspectratio]{figures/Margin2.pdf}
%     \end{center}
%   \caption{For the object instance in the middle of the left image, the \textbf{halo loss}  computes weighted cross-entropy between the mask shown on the right (red = positive constant, blue = negative constant) and each of the color channels (except the background channel). The value of the loss for the instance is then computed as the minimum over all object channels. The minimal loss is obtained by assigning all pixels to a single channel (value = 1), while ensuring that pixels in the halo region do not belong to this region (value=0). The instance can then be recovered as a connected component.}
% \label{fig:halo_region}
% \end{figure}

The goal of learning in our model is to ensure that for any training image each object instance is (a) colored using the same color (i.e.\ that all of its pixels will have high values in a certain output map $\Y[c]$ for some $c > 1$) and that  (b) pixels adjacent to such instance has low value in the same map. Under these two conditions, the post-processing stage will recover the object instance correctly as a connected component corresponding to color $c$. One consequence of this observations, is that the coloring network should assign two adjacent object instances to different output channels. Overall, this makes the training process of the network akin to graph coloring, where each object instance should be assigned to one of $C-1$ colors in a way that no adjacent instances are assigned to the same color.

%The way the desired behaviour is enforced is by using a new loss function that we call the \textit{halo loss}. The idea behind this loss is to ensure conditions (a) and (b) are satisfied for each object instance without explicitly assigning an object instance to a certain map. Avoiding such assignment prior to learning thus allows the training process to perform such assignment in a most natural way.

As mentioned above, we do not assign instances to colors in advance prior to learning. Instead, coloring is performed during loss computation and can be seen as a part of the loss. Let $\X$ be a training image that has $K$ object instances, and let $\M^k$ be the set of pixels contained within the $k$-th object instance in this image. We consider the \textit{halo region} $\Mn^k$ defined as the set of pixels that lie outside $\M^k$ within the margin distance $m$ from the pixels of $\M^k$. In other words, the set $\Mn^k$ is defined as the set difference of the morphological dilation of $\M^k$ in the image plane and $\M^k$:
\begin{equation}
    \Mn^k = \text{dilate}(\M^k,m)\;\; \backslash\;\; \M^k\,.
\end{equation}
Here, the margin distance $m$ is an important parameter and we discuss its choice later.

Let $\Y = \Psi(\X;\theta)$ be the output of the coloring network for the current value of the network parameters $\theta$. 
For the $k$-th object the coloring process then seeks the color that maximizes the following simple objective:
% \begin{equation} \label{eq:coloring}
%     c_k = \arg\max_{c=2}^C \left( \frac{1}{|\M^k|}\sum_{p \in \M^k} \log \Y[c,p] +
%     \mu \frac{1}{|\Mn^k|} \sum_{p \in \Mn^k}  \log (1-\Y[c,p]) \right)\,,
% \end{equation}
\begin{align} 
    c_k = & \arg\max_{c=2}^C ( \frac{1}{|\M^k|}\sum_{p \in \M^k} \log \Y[c,p] \nonumber \\
    & \qquad + \mu \frac{1}{|\Mn^k|} \sum_{p \in \Mn^k}  \log (1-\Y[c,p]))\,
    \label{eq:coloring}
\end{align}
In other words, the coloring selects the color in order to maximize the average log-probability of the color inside the object itself, and to minimize the average log-probability in the halo region. In \eq{coloring}, $\mu$ is another meta-parameter that controls the influence of the negative part. 

Once the instances are colored, we use the standard pixel-wise log-loss, treating $c_k$ as the pseudo ground-truth (during this particular epoch). Thus, the following loss is computed for each training image:
\begin{align} 
L(\X,\theta) = & -\sum_{k=1}^K \frac{1}{|\M^k|} \sum_{p \in \M^k} \log \Y[c_k,p] \nonumber \\
& \qquad - \sum_{p \in \text{Background}} \log \Y[1,p] \,
\label{eq:logloss}
\end{align}
and then back-propagated through the network, with the network parameters $\theta$ updated accordingly.

Overall, the learning process can be seen as the standard training of a semantic segmentaiton network, but with dynamically changing ``ground truth''. We have also tried a variant of the algorithm that uses the minus sum of the coloring criteria \eq{coloring} as the learning loss instead of the traditional log-loss \eq{logloss}, but have obtained inferior results.

Note that the coloring process discouraged the emergence of instances colored with the same color that are closer than the margin $m$ to each other. Thus, if at test-time two connected components of the same color are much closer than $m$, this presents a strong evidence that they correspond to the same object instance, which justifies the optional merging procedure.

\subsection{Meta-parameters and their influence}
\label{sect:params}

Like any other instance segmentation method that we are aware of, ours comes with a number of meta-parameters. Here, we discuss their choice and influence. We further evaluate the sensitivity of our method to these parameters in the experimental section.

\textbf{The number of colors $C$} is the easiest parameter to set. We have found that when provided an excessive number of maps, the algorithm automatically chooses not to use some of them. Thus, the number of maps should simply be chosen sufficiently high. Setting $C$ too small may lead to the method merging separate instances together.

\textbf{The size threshold $\tau$} allows to prune away small connected components, and thus controls the trade-off between false positives and false negatives. The parameter is set through validation, which is computationally ``cheap'' since trying different thresholds does not require retraining the coloring network.  %In general, we have found that it is better to have a separate threshold $\tau_c$ for for each of the $C-1$ non-background colors. This is because, as shown below, the network often learns to use different colors for instances of different sizes, and therefore a single size threshold may be suboptimal. We have investigated two different schemes for setting $\tau_c$. In the simpler scheme, we run the learned network on the validation set (TODO:CHECK), find all instances that have been colored with the color $c$, then simply set $\tau_c$ to the size of the smallest connected component corresponding to these objects. Alternatively, we pick optimal $\tau_c$ in order to maximize the performance score using search the validation set. Note that none of the two variants are time-consuming, since trying different thresholds does not require rerunning the network training.

\textbf{The margin $m$ and the halo weight $\mu$} are also set through the validation procedure. Generally, they control the trade-off between fragmentation and undersegmentation. Thus, if $m$ or $\mu$ are too small, the learning process will mostly focus on coloring all pixels of the same instance with the same color, and will care less about having this color distinct from the remaining objects, which can lead to undersegmentation (using same color for adjacent instances). Having $m$ and $\mu$ large leads to the learning process caring mostly about spreading pixels from the nearby instances into different maps, which may lead to their fragmentation. 

\textbf{The merging threshold $\rho$.} Merging is only used if partial occlusions are common in the dataset. E.g.\ it is needed for autonomous driving, where vehicles are occluded by pedestrians and posts, but is not needed for instance segmentation tasks associated with monolayer cell cultures. When merging is performed, $\rho$ controls the trade-off between fragmentation (low $\rho$, sometimes parts of the same instance not merged) and undersegmentation (large $\rho$, sometimes different instances are merged). As in the case with the size threshold, trying different merging thresholds on the validation set is cheap since it does not require retraining the coloring network.

\section{Experiments}\label{sec:experiments}
%\section{Experiments}

We evaluate our approach on three datasets, including two standard benchmarks for instance segmentation: the CVPPP plant phenotyping dataset \cite{scharr2014annotated} and the CityScapes dataset \cite{cordts2016cityscapes}. The third dataset is composed of bright-field microscopy images of E.Coli organisms.
The experiments were performed on a NVidia Titan X GPU. The training in all cases was performed using ADAM optimizer with the learning rate 1e-3. All parameters for our method were evaluated on the validation sets, while the ground truth for the CVPPP and CityScapes challenges is withhold.

\begin{figure*}
\begin{center}
\includegraphics[width=\linewidth,keepaspectratio]{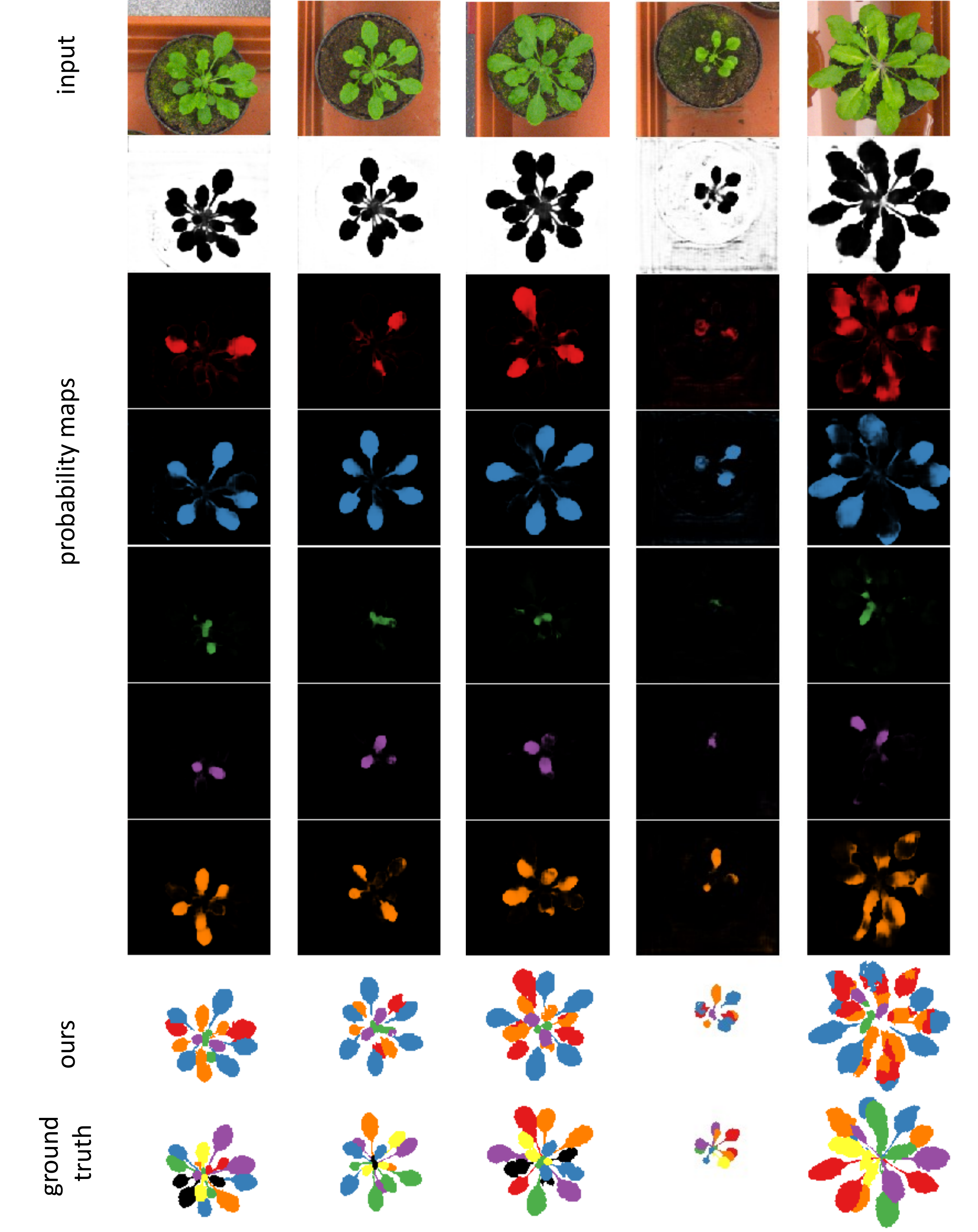}
\end{center}
   \caption{Representative results of the six-color model on the validation set of the CVPPP dataset. For the input (top), the coloring network colors pixels into six colors (next six rows), after which the connected component analysis recovers the instances that closely match the ground truth (bottom).}
\label{fig:leaves}
\end{figure*}

\subsection{CVPPP dataset}

The Computer Vision Problems in Plants Phenotyping (CVPPP) dataset~\cite{minervini2016finely} is the one of the most popular instance segmentation benchmarks. The images in this dataset are challenging because of complex occlusions between leafs, high variety of leafs shapes and backgrounds. The training set consists of 128 top-down view images, with $530 \times 500$ pixels size each and a hidden testset that includes 33 images. The accuracy of the instance segmentation is estimated using \textit{symmetric best Dice} coefficient (\textit{SBD}) (c.f.~\cite{scharr2016leaf}). The second parameter, which is less directly related to instance segmentation, is the absolute difference in counting $\mid$DiC$\mid$. 

To accelerate the training procedure we down-scaled each image to the size ($265 \times 250$ pixels). Due to relative small number of annotated images the training process requires augmentation. We applied three random transforms: cropping of patches $192 \times 192$ pixels, rotation to a random angle and flips along axes. The batch size was fixed to 40 patches.
In these experiments, we stuck to a relatively small UNet-type architecture~\cite{ronneberger2015u} consisting of 4 downscaling (convolution+ReLU) and upscaling (transposed convolution+ReLU) modules with skip connections.%Each downscale module consist of sequence convolutions with ReLUs non-linearity. The length of this sequence was proportional to the scale level. The upscaling modules are simply transposed convolution followed by ReLU. 

To retrieve leaves instances from the output channels the argmax operation is performed. At this point for each pixel we have a channels label. The connected component algorithm is applied and components smaller that a specified size are removed. The thresholds were selected by maximization of the average SBD value on the validation set.

\begin{table}
    \begin{center}
        \begin{tabular}{|l|c c|}
        \hline
        \textit{Method} &$\mid DiC\mid$&$SBD(\%)$\\
        \hline
        IPK \cite{pape20143}&2.6&74.4\\
        Nottingham \cite{scharr2016leaf}&3.8&68.3\\
        MSU \cite{scharr2016leaf}&2.3&66.7\\
        Wageningen \cite{yin2014multi}&2.2&71.1\\
        PRIAn \cite{giuffrida2016learning}&1.3&-\\
        Recurrent IS \cite{romera2016recurrent}&1.1&56.8\\
        Recurrent IS+CRF \cite{romera2016recurrent}&1.1&66.6\\
        Recurrent with attention \cite{ren2016end}&0.8&\textbf{84.9}\\
        Embedding-based \cite{de2017semantic}&1.0&84.2\\
        \hline
        \textbf{Deep coloring}&2.0&80.4\\
        \hline
        \end{tabular}
    \end{center}

    \caption{Benchmark results on the CVPPP dataset, where deep coloring was trained end-to-end and evaluated without the use of provided masks (which explains part of the overcounting). The Symmetric Best Dice coefficient ($SBD$) reflects the instance segmentation accuracy.}
    \label{tab:cvppp}
\end{table}

The main network has nine output color channels (including background) and was trained for 20000 iterations with margin $m$ equal to 21 pixels and $\mu$ equal to seven (selected on the validation set, which happens to be approximately equal to the average diameter of the object in training set). We have also trained a version with six output channels (which performed worse by about 1\% SBD on the validation set) and used it for the visualization of the results.  

The results of the nine-channel deep coloring architecture as well the results of other methods evaluated on the dataset are in \tab{cvppp}. Our method performs better than most methods evaluated on this dataset except for the work \cite{de2017semantic} and the attention-based approach \cite{ren2016end}, which is arguably considerably more complex. We do not use the ground truth background masks provided for the evaluation (used by most other methods including \cite{de2017semantic}). Instead, we allowed our method to perform background segmentation simultaneously with performing instance segmentation. 

In order to make a more direct comparison with the related work~\cite{de2017semantic}, we took its reimplementation, for which we use the U-Net backbone architecture (exactly same as ours). The dimensionality of the output for ~\cite{de2017semantic} was set to 8 per pixel (same as recommended in \cite{de2017semantic}). A full mean-shift procedure was used as postprocessing (\texttt{sklearn} implementation). The bandwidth of meanshift was tuned using validation set. On the CVPPP dataset we found that the method reached \textbf{0.84} SBD on the validation set for the optimal bandwidth, while dropping sharply when the bandwidth was different (0.78 SBD for 0.8x optimal bandwidth and 0.79 SBD for 1.2x optimal bandwidth).
Our approach reaches \textbf{0.87} SBD on the validation set in the same protocol (with the size threshold tuned on the validation). We have also evaluated the oracle variant of \cite{de2017semantic} where mean-shift is replaced with k-means, whereas the number $k$ of instances is taken from the ground truth. This variant reached 0.91 SBD. Our conclusion is that setting the parameters of the connected component analysis (minimal size threshold) inherent in our method is easier than setting the meanshift bandwidth inherent to \cite{de2017semantic}. We also observe that our method trains faster (about $2$x till convergence). Our postprocessing is orders of magnitude faster than sklearn-meanshift ($0.05$s vs $30$s). Compared to the fast mean shift version used in \cite{de2017semantic}, the operation complexity of our postprocessing is still lower.

% \begin{figure}
% \begin{center}
% \includegraphics[width=\linewidth,keepaspectratio]{figures/Examples2.pdf}
% \end{center}
%   \caption{To probe the model, we show the examples of leaves (in the center of the image) clustered together by our loss by being colored with the same color during training (one row per color).    The clustering seems to depend on the individual and relative properties of each instance.}
% \label{fig:clusters}
% \end{figure}

Sample results of our model (for the six-color variant) are given in \fig{leaves}. In general, the system successfully learns both to discern the instances and to segment background. The main failures are on the atypical plants with large leaves, which have very few examples in the training set. It is interesting to observe the clustering behaviour as leaves of similar type (in terms of size and location) have been colored consistently into the same color across different images.

To assess the parameter sensitivity, we have estimated the dependency of the SBD error measure on the negative part influence meta-parameter $\mu$ (vertical) and the margin/halo size (horizontal) on the CVPPP validation set (\fig{paramsens}). The number of maps were redundantly fixed to nine in all our experiments, with the model typically using a subset of those. Additionally, we also provide the results for the ResNet-34 architecture \cite{he2016deep}. We have observed that ResNet-34 achieves worse results at least for the same number of epochs as was used to train the U-Net architecture. Training the ResNet-based architecture for more epochs recovered part of the gap between the two architectures.

\if@twocolumn%
\begin{figure}
\begin{center}
\begin{tabular}{cc}
\includegraphics[height=4cm,keepaspectratio]{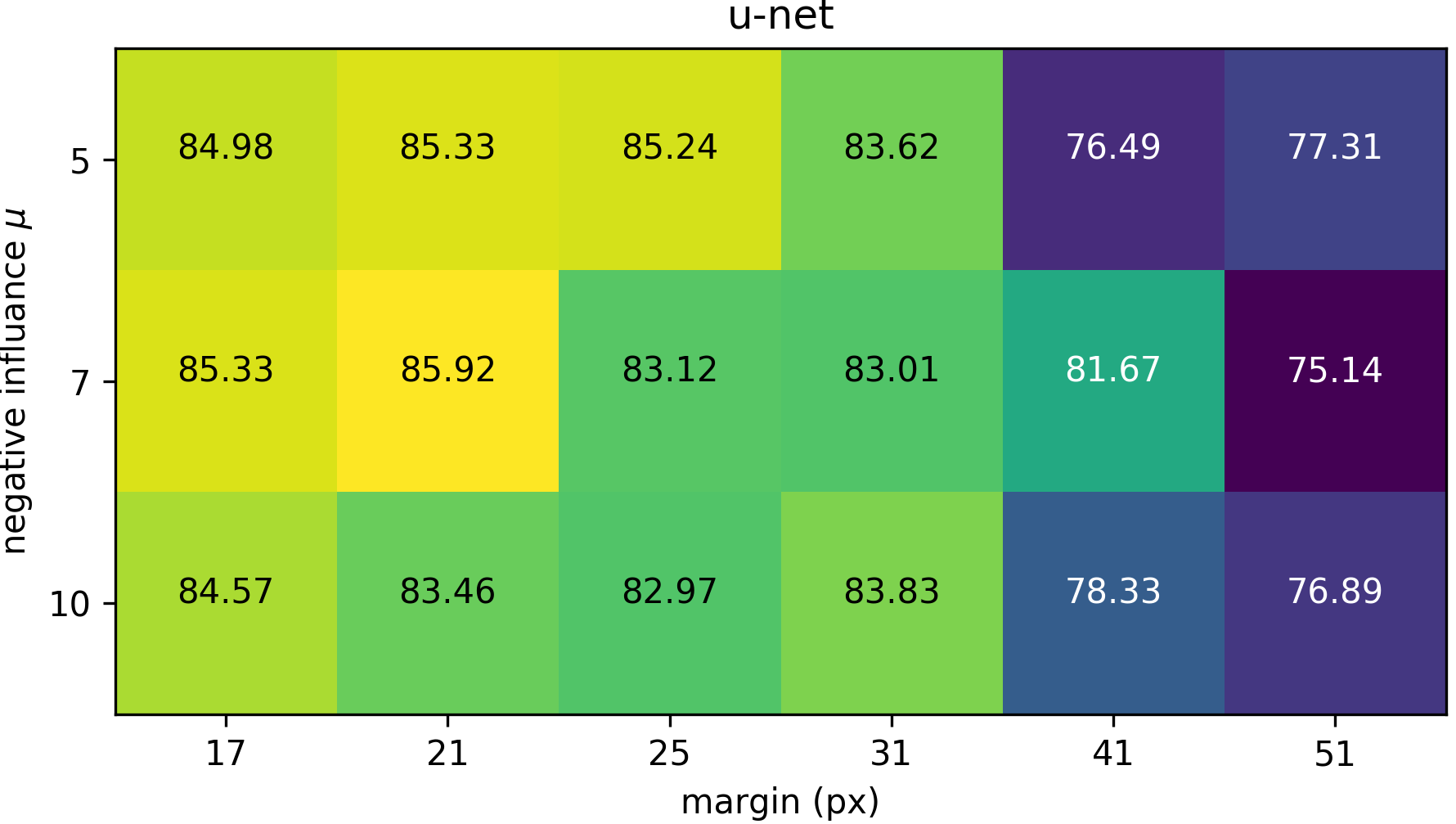}\quad&\quad
\includegraphics[height=4cm,keepaspectratio]{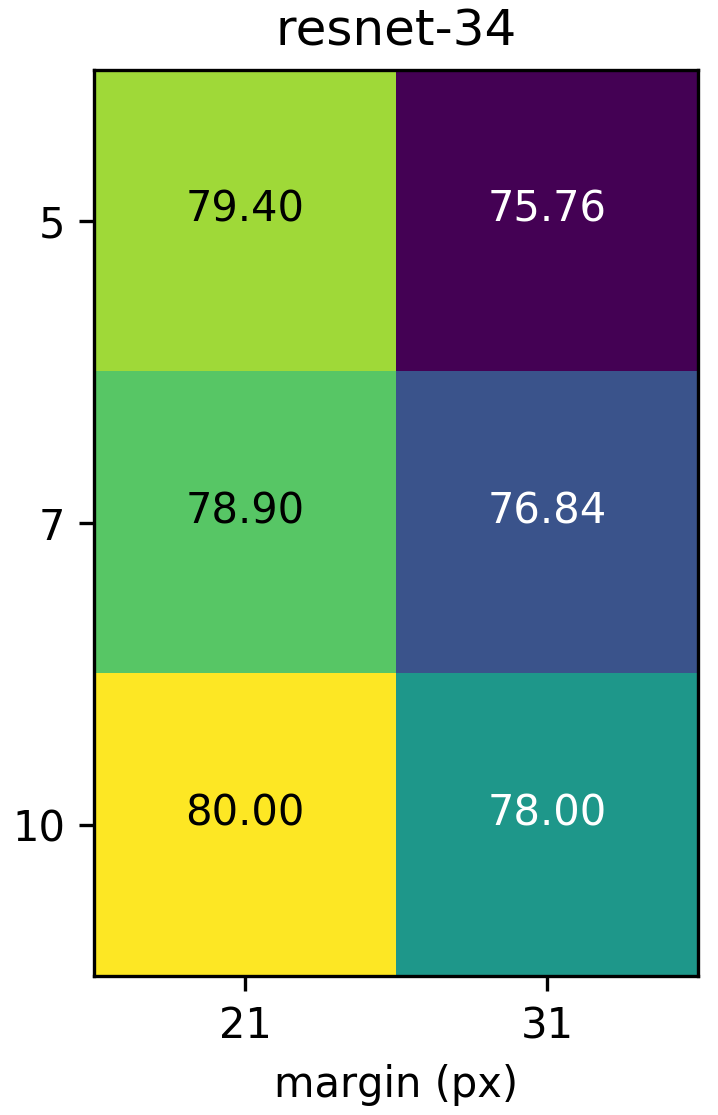}
\end{tabular}
\end{center}
  \caption{Parameter sensitivity analysis on the CVPPP validation test for two different network architectures. Generally, our approach is not overly sensitive to considerable changes of the margin size and the $\mu$-parameter around the optimum.}
\label{fig:paramsens}
\end{figure}
\else% \@twocolumnfalse
\begin{figure}
\begin{center}
\begin{tabular}{cc}
\includegraphics[height=3.5cm,keepaspectratio]{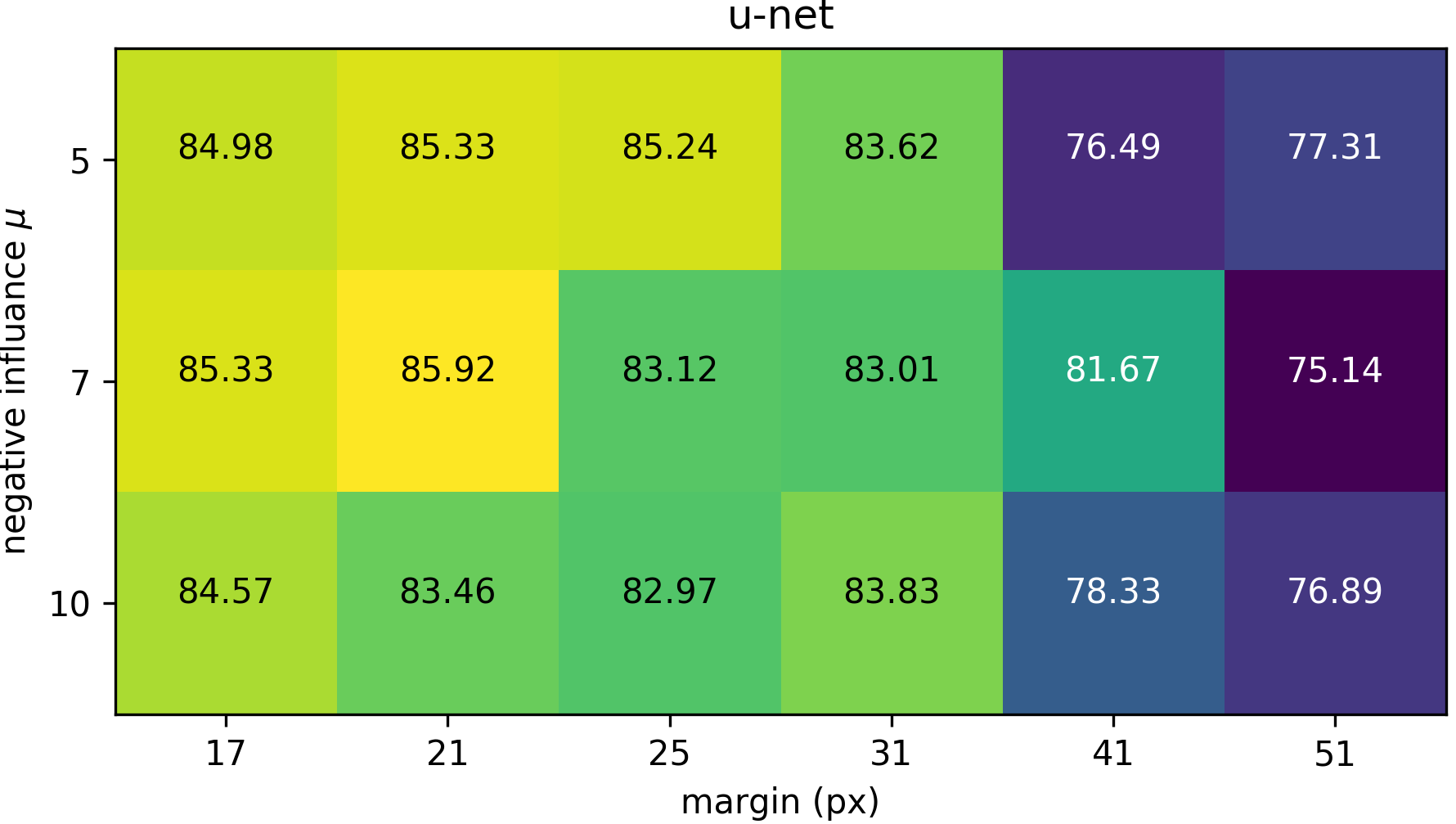}&
\includegraphics[height=3.5cm,keepaspectratio]{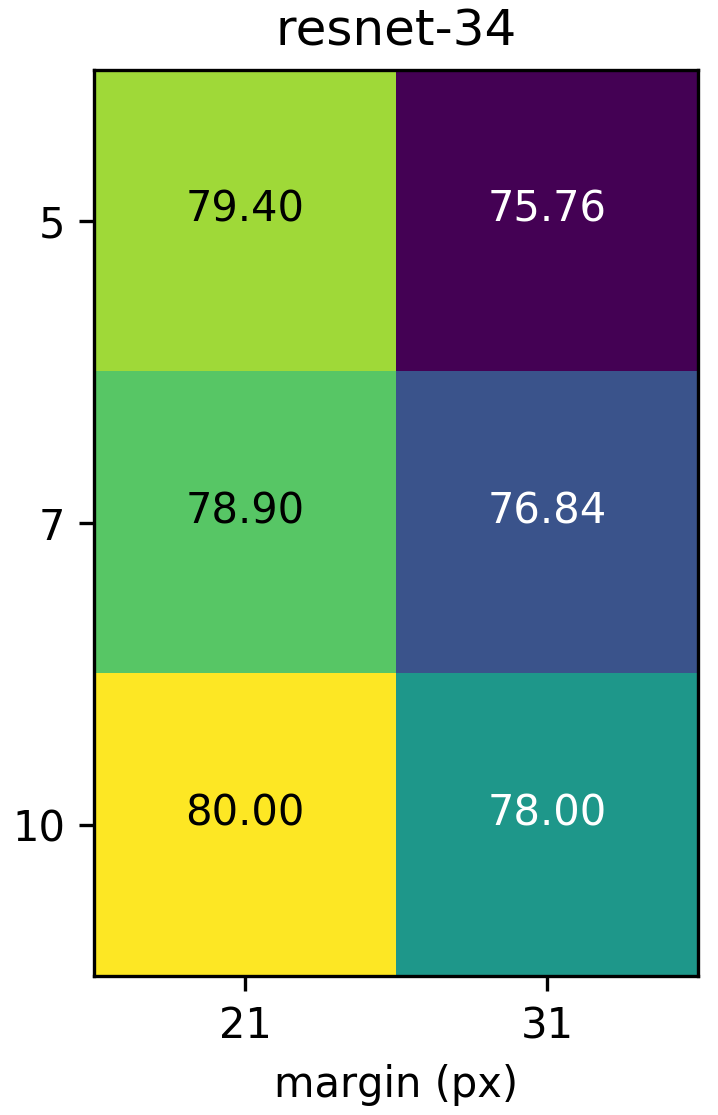}
\end{tabular}
\end{center}
  \caption{Parameter sensitivity analysis on the CVPPP validation test for two different network architectures. Generally, our approach is not overly sensitive to considerable changes of the margin size and the $\mu$-parameter around the optimum.}
\label{fig:paramsens}
\end{figure}
\fi
 
% \begin{figure}
% \begin{center}
% \begin{tabular}{cc}
% \includegraphics[width=2cm,height=\textheight,keepaspectratio]{figures/u-net.png}\quad&\quad
% \includegraphics[width=2cm,height=\textheight,keepaspectratio]{figures/resnet_34.png}
% \end{tabular}
% \end{center}
%   \caption{Parameter sensitivity analysis on the CVPPP validation test for two different network architectures. Generally, our approach is not overly sensitive to considerable changes of the margin size and the $\mu$-parameter around the optimum.}
% \label{fig:paramsens}
% \end{figure}

\subsection{Microscopy image dataset}

\begin{figure}
\begin{center}
\includegraphics[width=\linewidth,keepaspectratio]{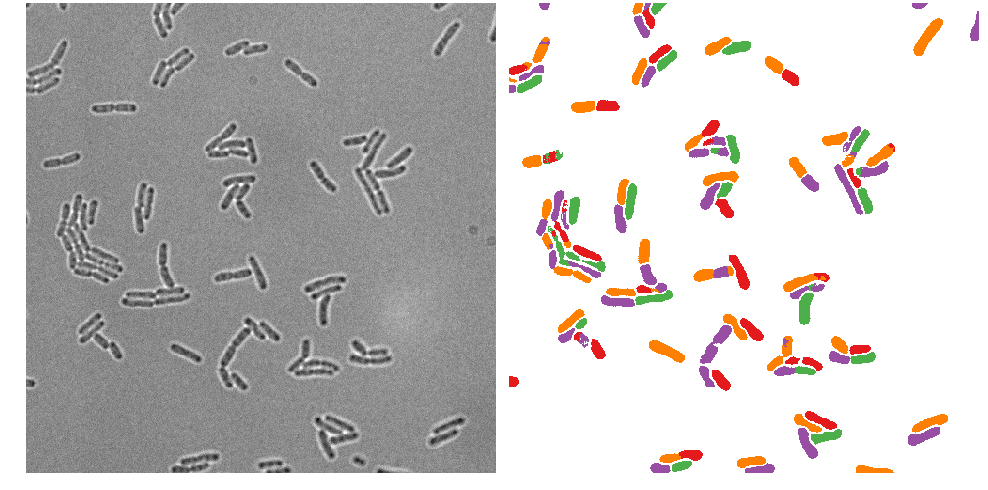}
\includegraphics[width=\linewidth,keepaspectratio]{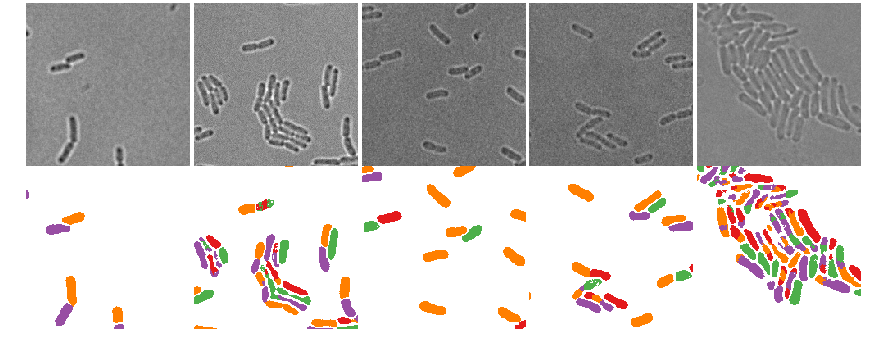}
\end{center}
  \caption{Representative results on the microscopy image dataset. Our approach can handle large number of organisms gracefully. In the result, the color used for an organism can be determined both by individual properties (e.g.\ long organisms are often colored red) as well as the participation in a certain multi-organism pattern (note multiple orange-magenta pairs).}
\label{fig:ecoli}
\end{figure}

We then present the results on the dataset of microscopy images of E.Coli organisms, corresponding to individual instances \fig{ecoli}. The dataset is hard for two reasons. First, the number of organisms is large and they are crowded. Secondly, the organisms divide by splitting in the middle. The splitting process results in a subtle change of the appearance, and there is a very small visual difference between the appearance of the parent organism just before splitting and the appearance of the two daughter organisms immediately after splitting. Additionally, the ground truth is derived from the annotation that is composed of line segments (one line segment per organism) and is approximate.

We have run the approach with the network $\Phi$ having the same U-Net architecture as for the CVPPP experiments. We have observed that for this dataset the training process chose only five colors to do the coloring.

As a baseline, we run the U-Net semantic segmentation \cite{ronneberger2015u} with the same network architecture and three output labels (`background', `interior', `boundary'). The instances were identified as connected components of the interior class. As suggested in \cite{ronneberger2015u} we increased the weight of the 'boundary' class, while tuning it on the validation set. The result of this study is mean SBD for U-net is \textbf{59.3\%}, the proposed method achieve \textbf{61.9\%} using the same architecture. In this case, we were not able to tune the method \cite{de2017semantic} to work well at all (the highest SBD that we were able to achieve was 44\%). Even the oracle version (k-means with prespecified number of instances) was able to achieve SBD 53\% only. While our failure is not conclusive, it seems that using clustering in medium dimensional space to identify instances (as utilized in \cite{de2017semantic}) becomes much harder for problems with very large number of instances, whereas the connected component analysis (as utilized in our method) scales more gracefully.

%	We perform two types of experiments, a comparative study of our method in the task of instance segmentation on popular datasets and a quantitative study of the hyperparameters influence. Before experiment description we present common experiment settings and implementation details.
	
% 	\subsection{Experimental setup and implementation details}

	%Иплементация
% 	We have implemented our approach using two deep learning frameworks: PyTorch~\cite{pytorch} and TensorFlow~\cite{tensorflow2015-whitepaper}. The source code and pretrained models will be made publicly available at the time of publication. All experiments were performed on a single NVidia TitanX GPU.
	
	%Оптимизатор и его параметры
% 	We train the whole network in an end-to-end manner using Adam optimizer with learning rate $10^{-3}$. Our training procedure was split into two stages: the annihilating stage and the refinement stage. 
	
	%Постпроцессинг
%	In order to compare our result we need to provide binary masks for each instance. This is done in two steps: building a discrete index map and detection of connected components. The index map is a discrete image where each pixel value is associated with an instance group. To build an index map from probability maps we use pixel-wise $argmax$ function. A connected components detector is used to retrieve individual instances on the index map. To prevent appearance of noisy instances we applied a thresholds on instance size. The value of the thresholds are selected to minimize the error on the training set.

\newlength{\lennine}
\setlength{\lennine}{160pt}

\begin{figure*}
\begin{center}
\setlength\tabcolsep{2pt}
\begin{tabular}{ccc}
     input & ours & ground truth \\
     
        \setlength{\fboxsep}{0pt}      
        \fbox{\includegraphics[width=\lennine,keepaspectratio]{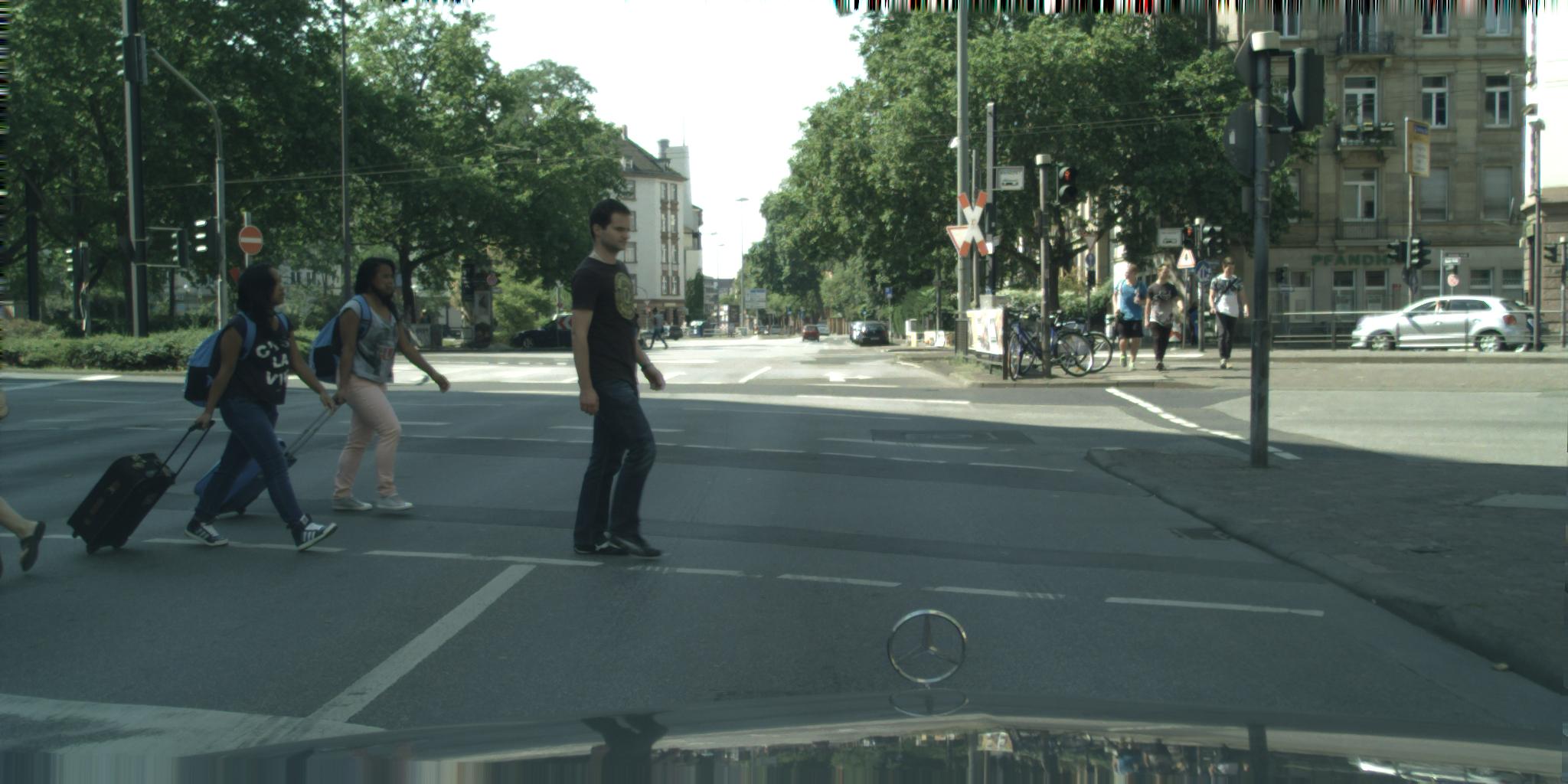}} & 
        \setlength{\fboxsep}{0pt}         \fbox{\includegraphics[width=\lennine,keepaspectratio]{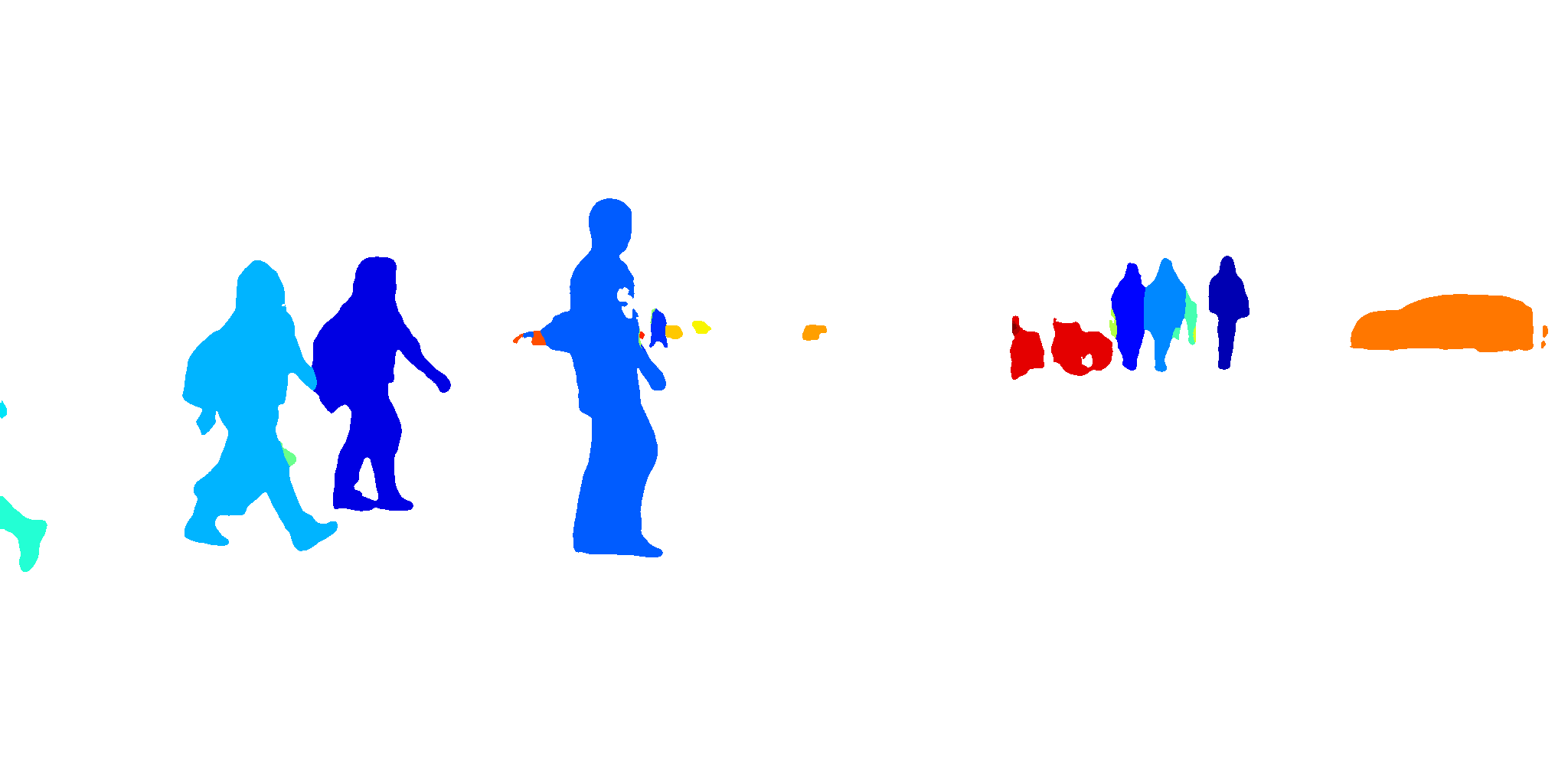}} & 
        \setlength{\fboxsep}{0pt}         \fbox{\includegraphics[width=\lennine,keepaspectratio]{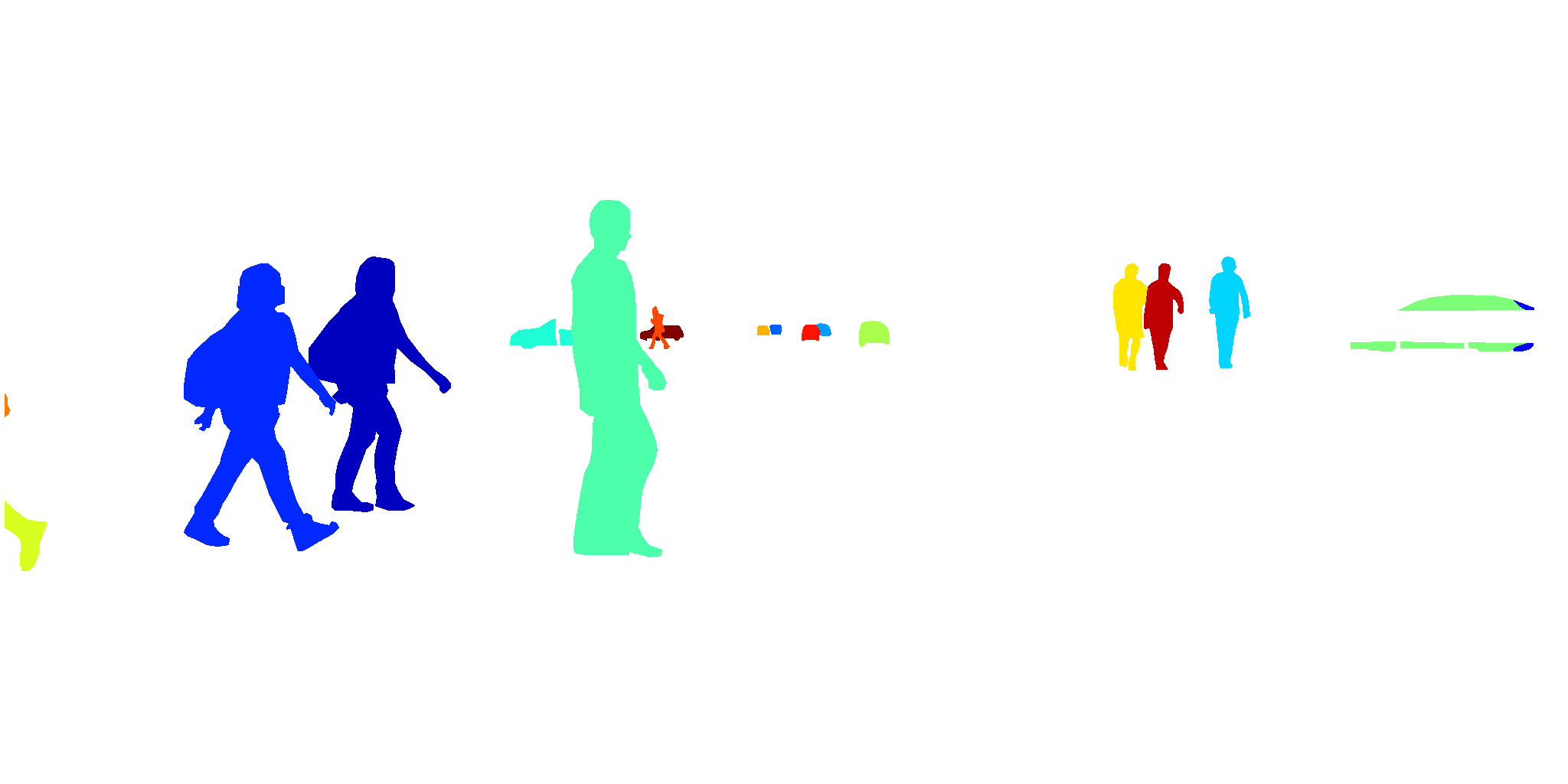}}
    \\
        \setlength{\fboxsep}{0pt}     \fbox{\includegraphics[width=\lennine,keepaspectratio]{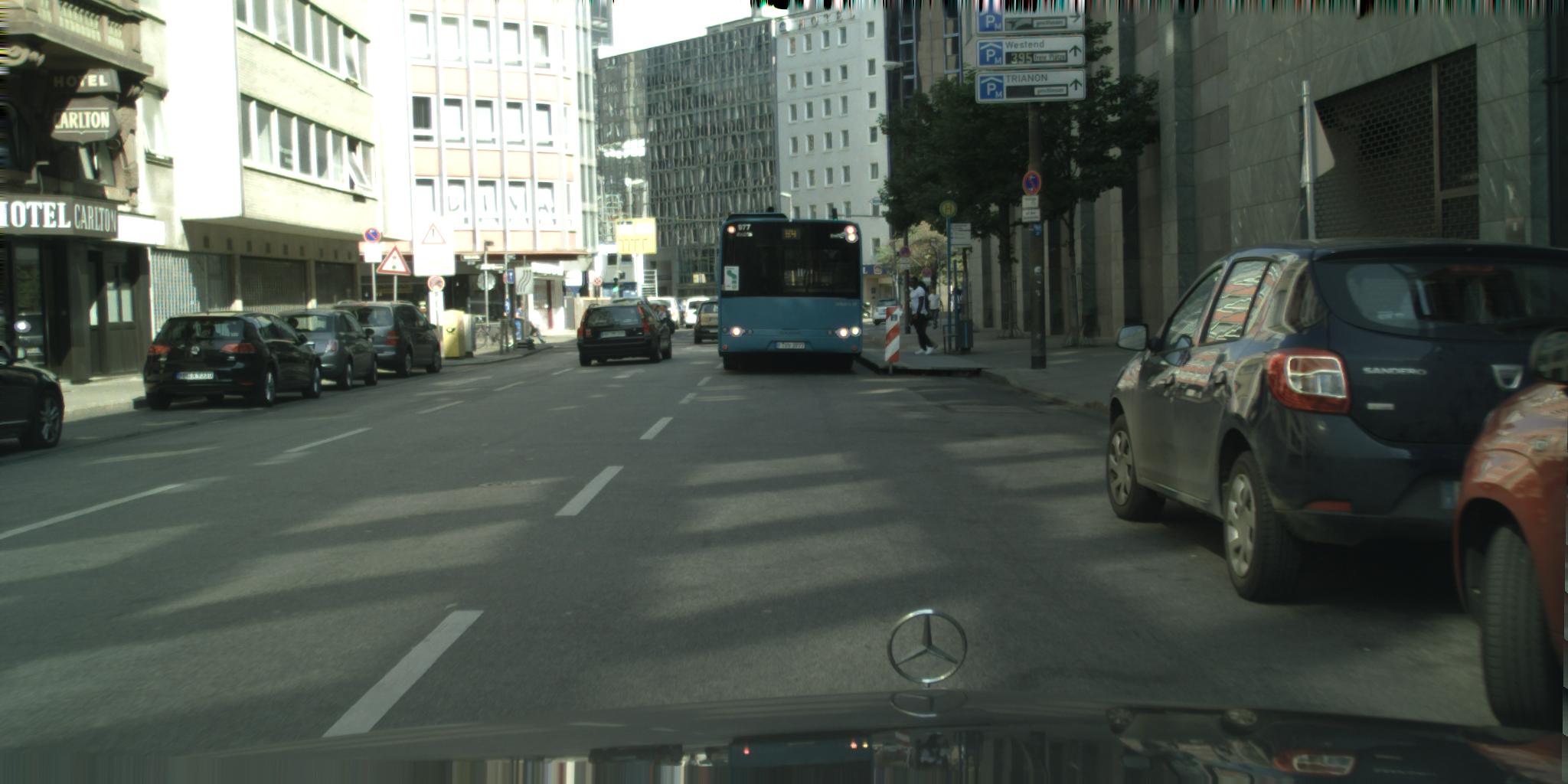}} &
        \setlength{\fboxsep}{0pt}  \fbox{\includegraphics[width=\lennine,keepaspectratio]{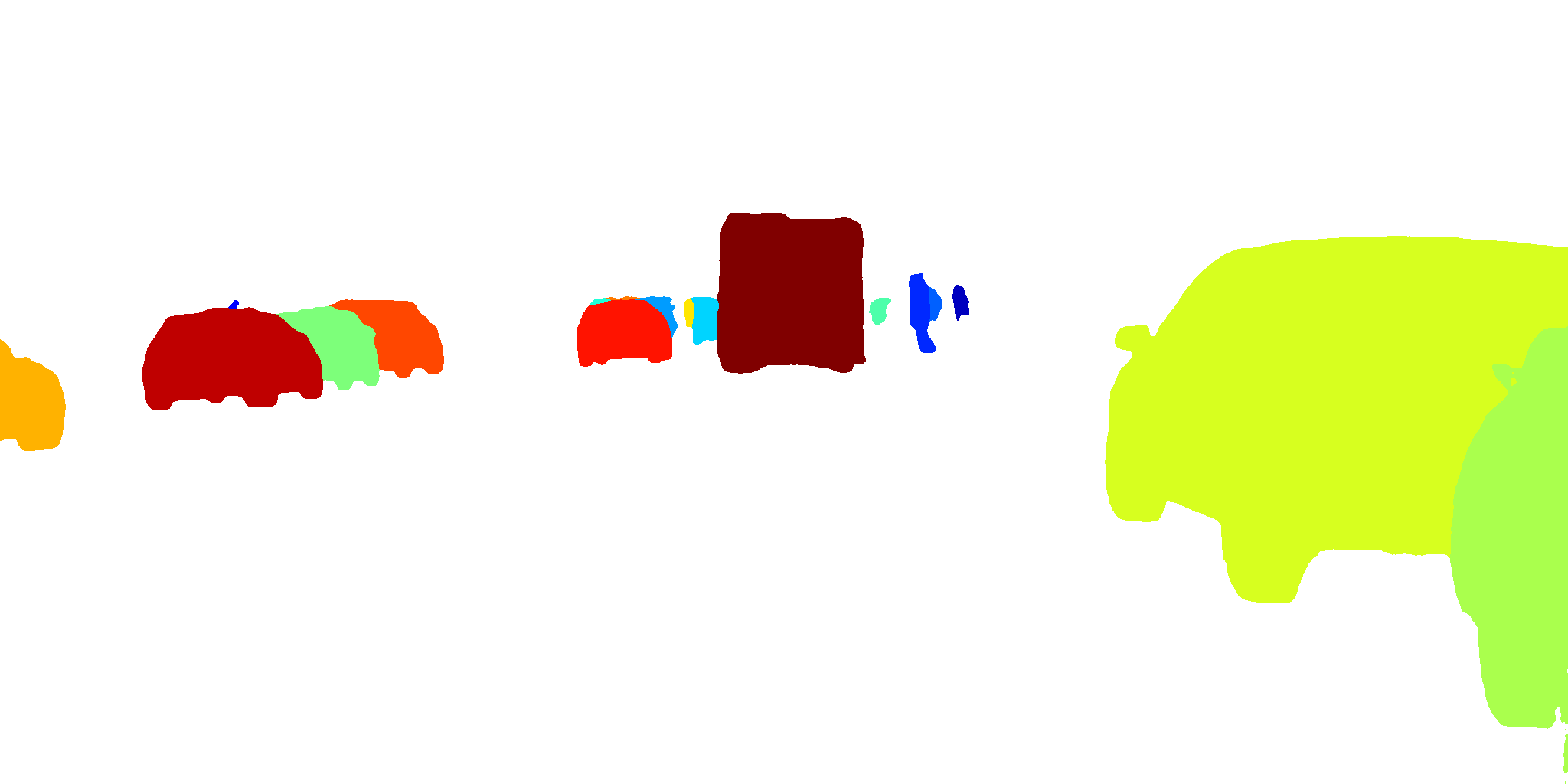}} & 
        \setlength{\fboxsep}{0pt}  \fbox{\includegraphics[width=\lennine,keepaspectratio]{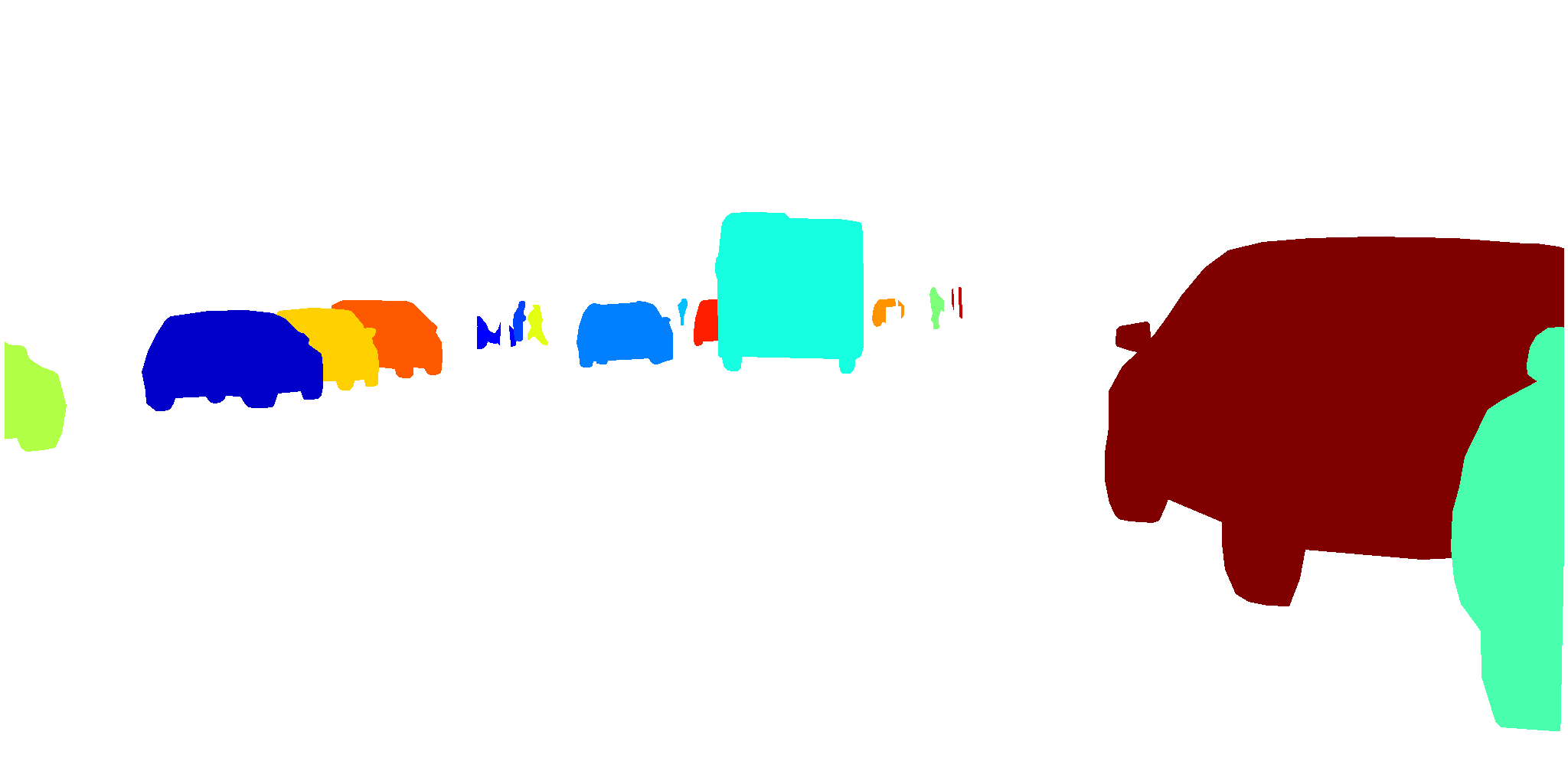}}
    \\
        \setlength{\fboxsep}{0pt}     \fbox{\includegraphics[width=\lennine,keepaspectratio]{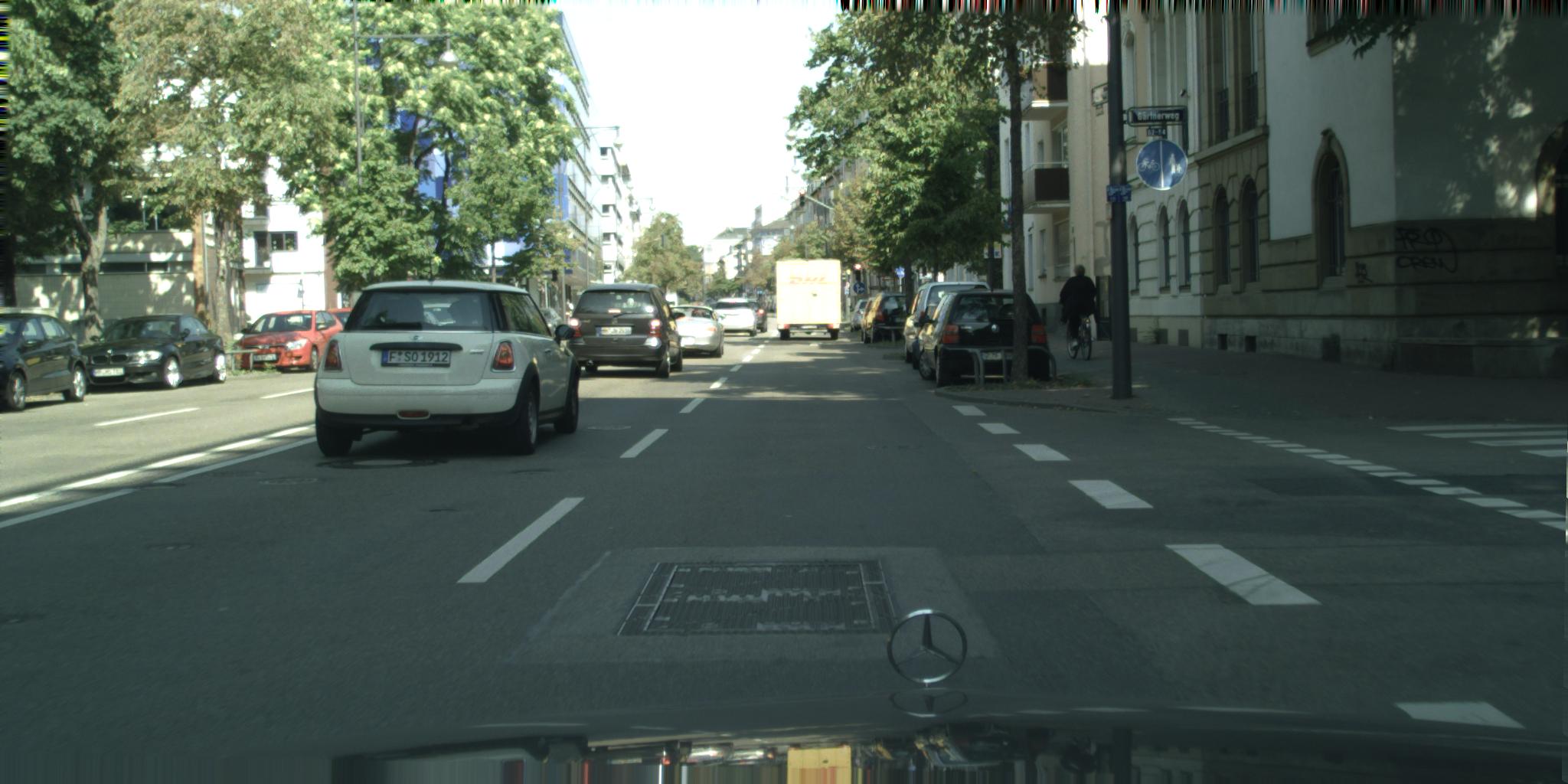}} &
        \setlength{\fboxsep}{0pt}  \fbox{\includegraphics[width=\lennine,keepaspectratio]{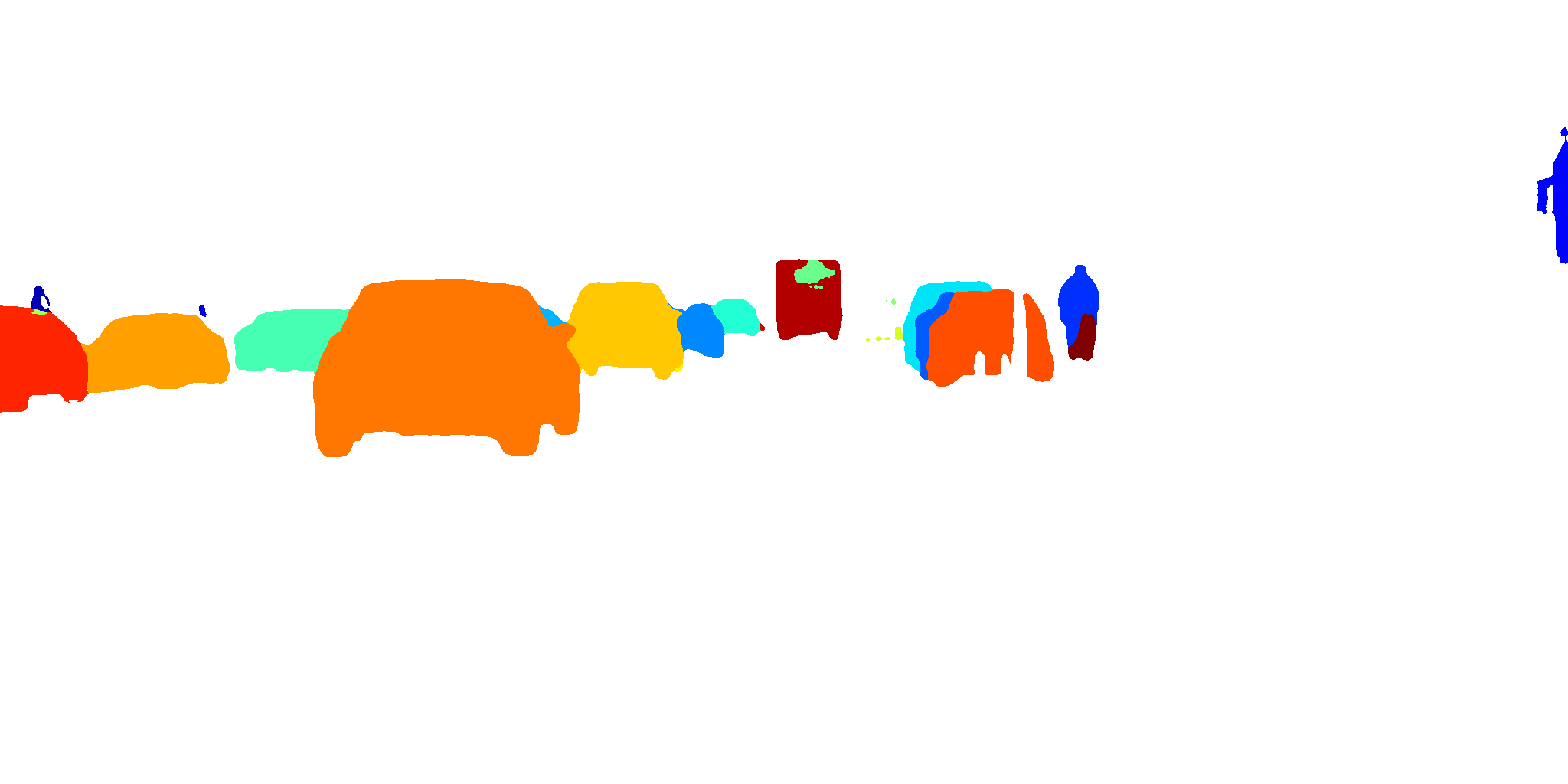}} & 
        \setlength{\fboxsep}{0pt}  \fbox{\includegraphics[width=\lennine,keepaspectratio]{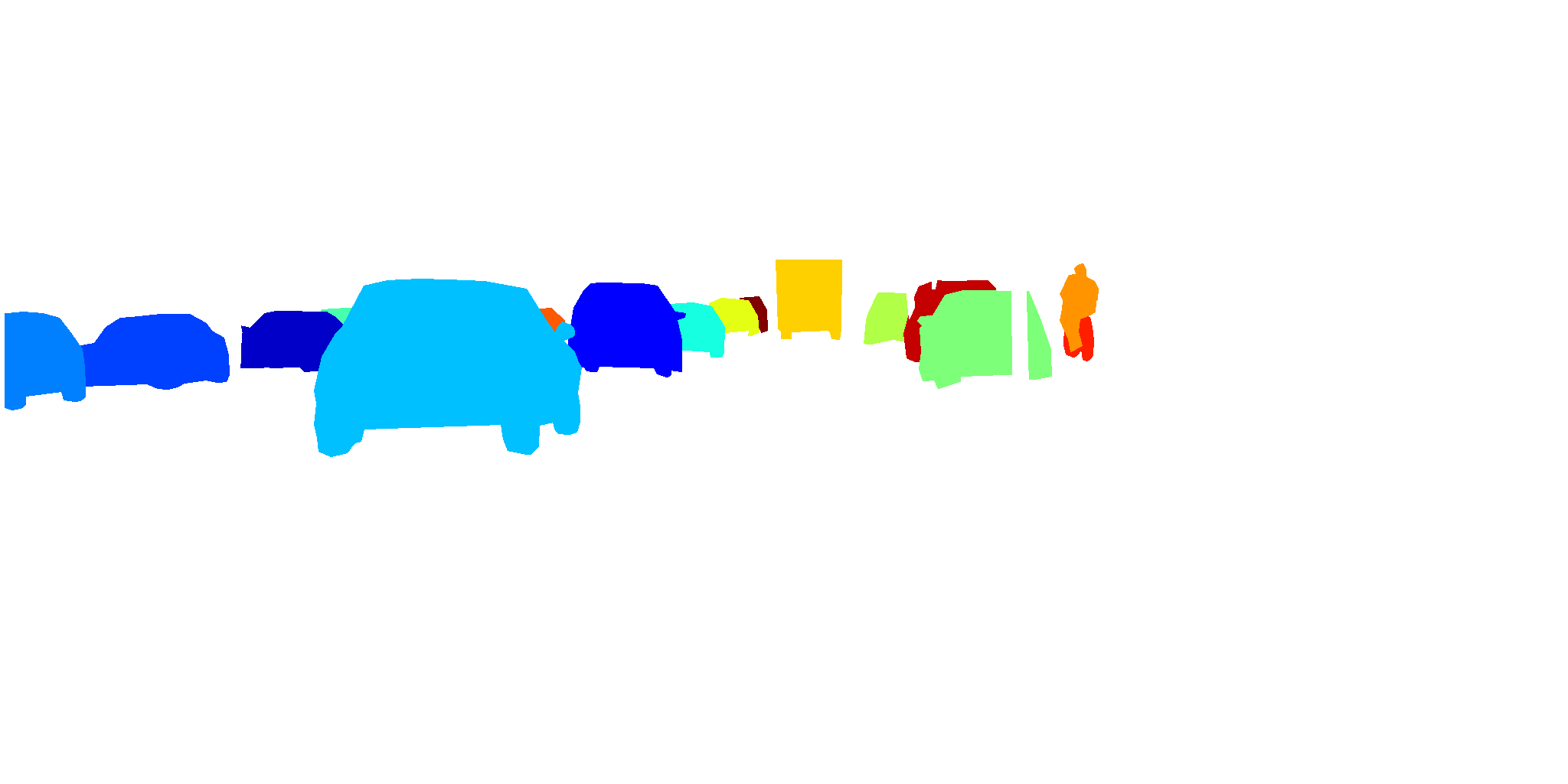}}
    \\
        \setlength{\fboxsep}{0pt}     \fbox{\includegraphics[width=\lennine,keepaspectratio]{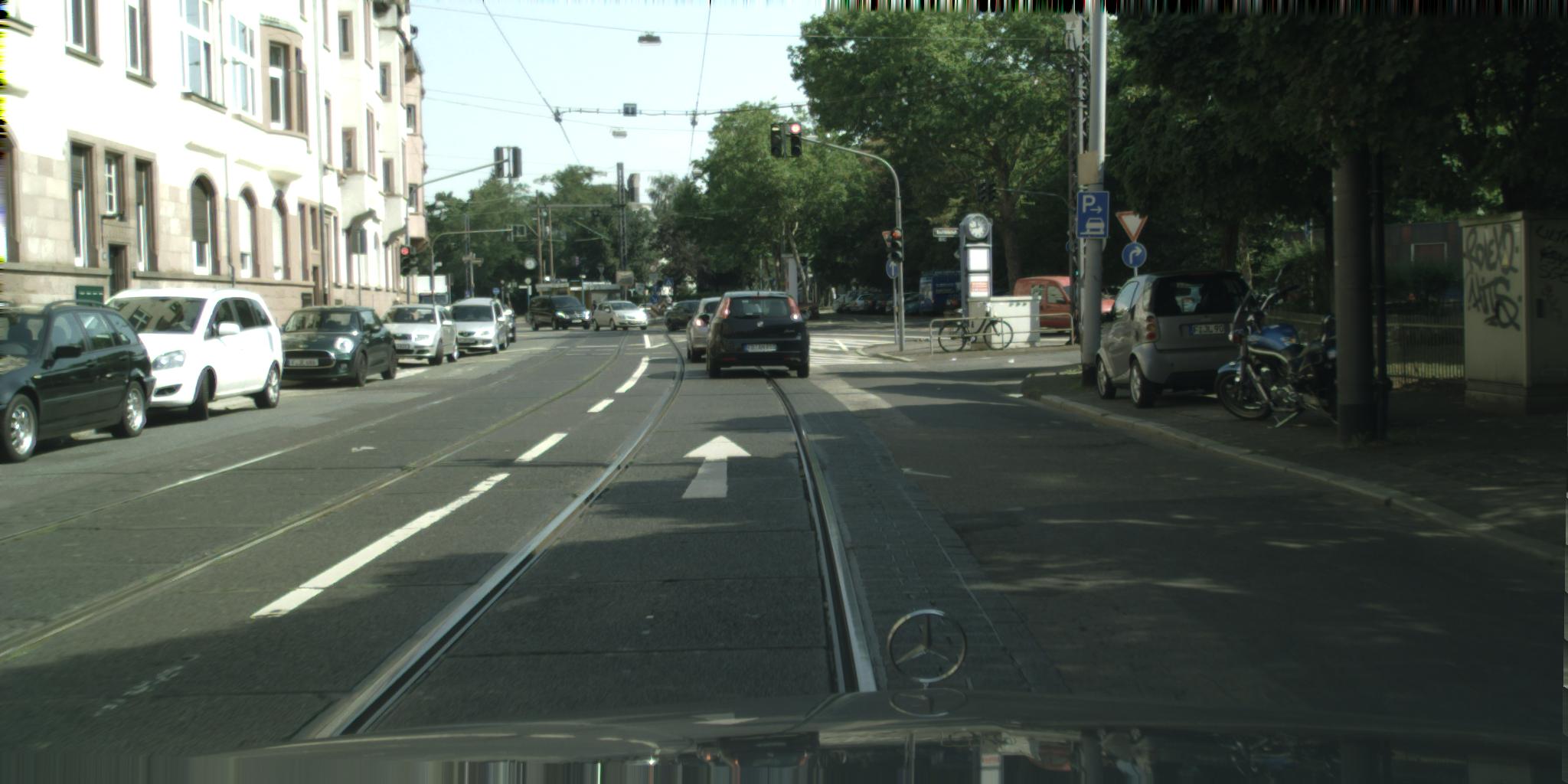}} &
        \setlength{\fboxsep}{0pt}  \fbox{\includegraphics[width=\lennine,keepaspectratio]{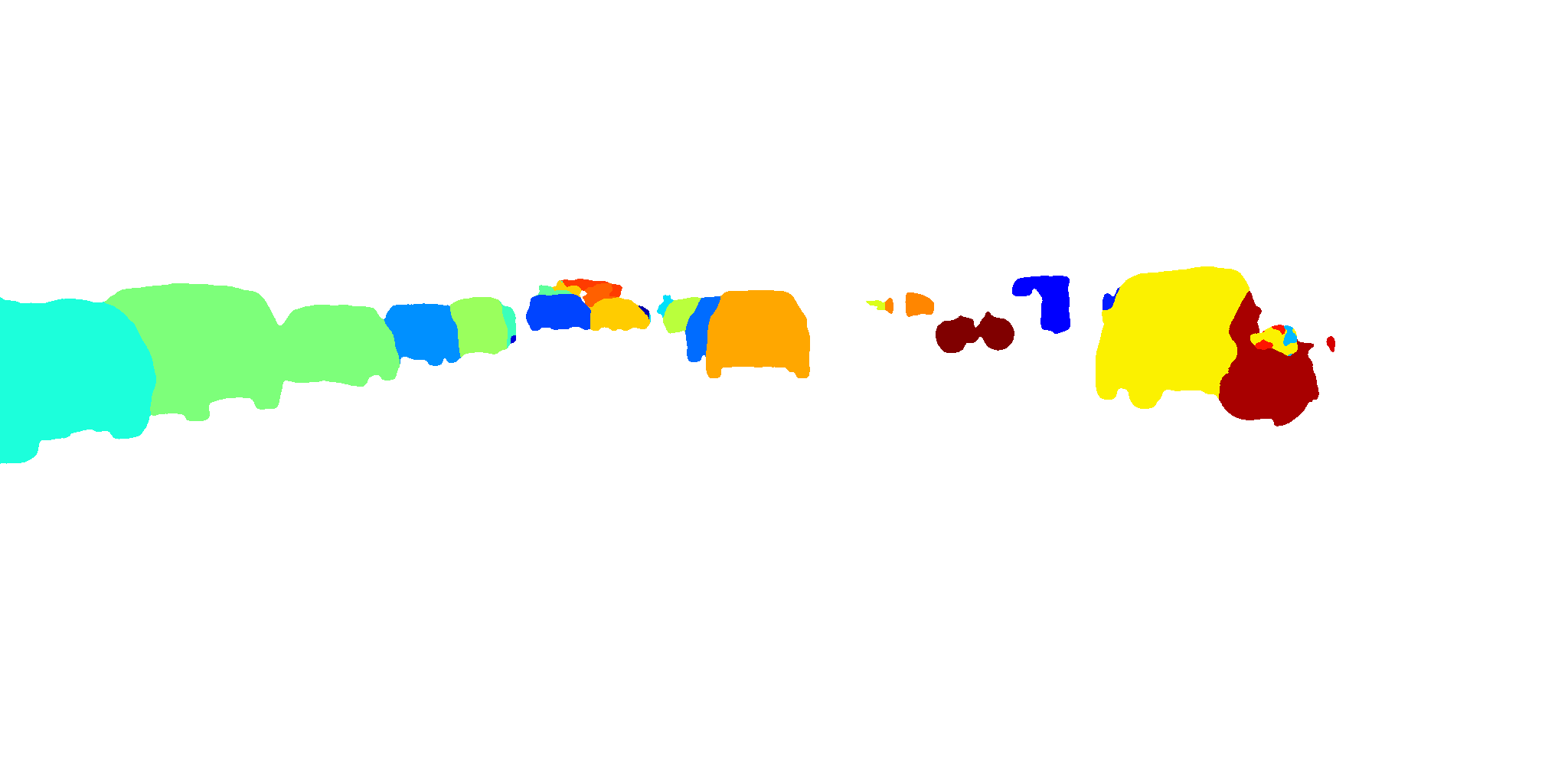}} & 
        \setlength{\fboxsep}{0pt}  \fbox{\includegraphics[width=\lennine,keepaspectratio]{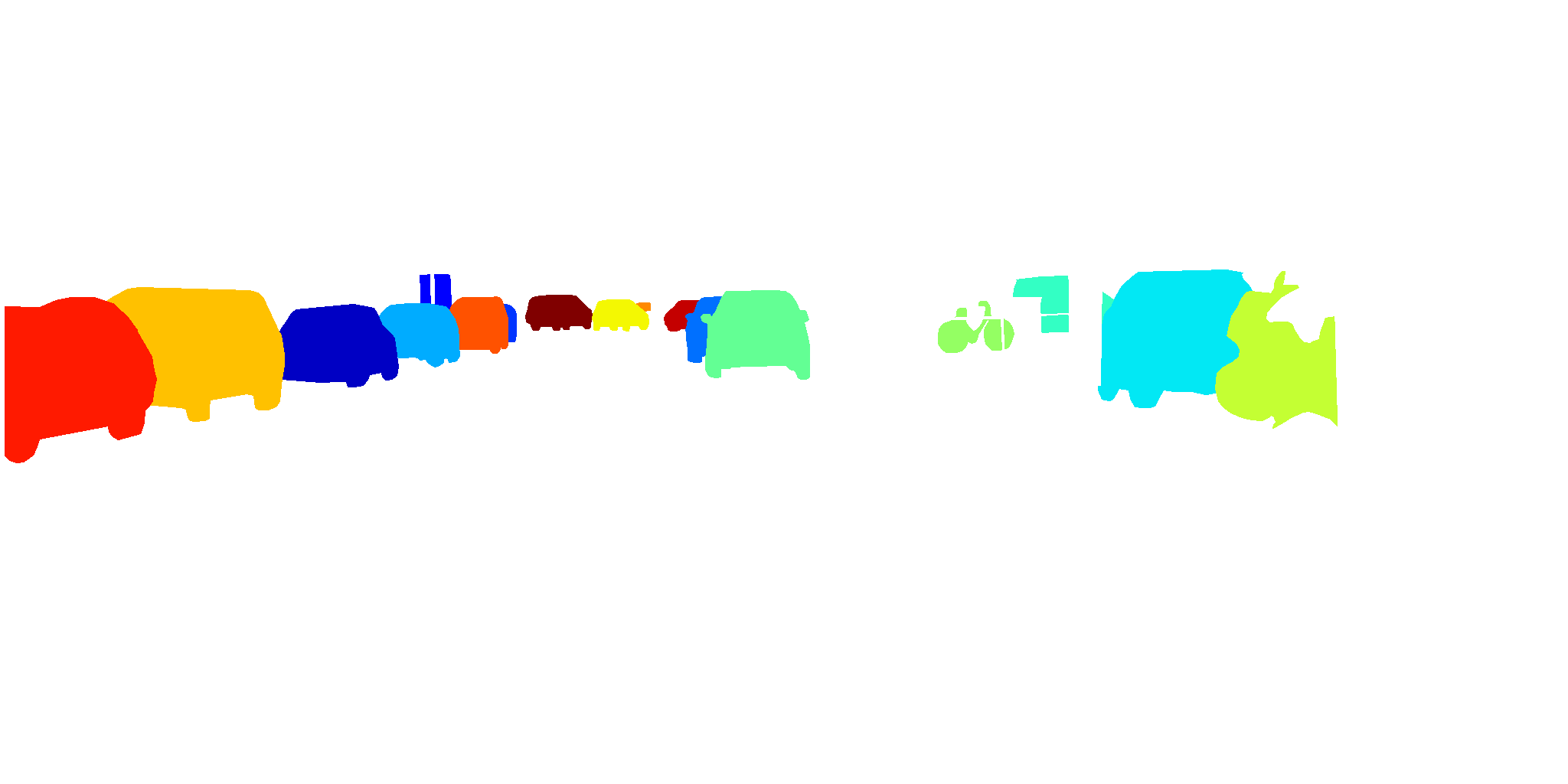}}
    \\
        \setlength{\fboxsep}{0pt}     \fbox{\includegraphics[width=\lennine,keepaspectratio]{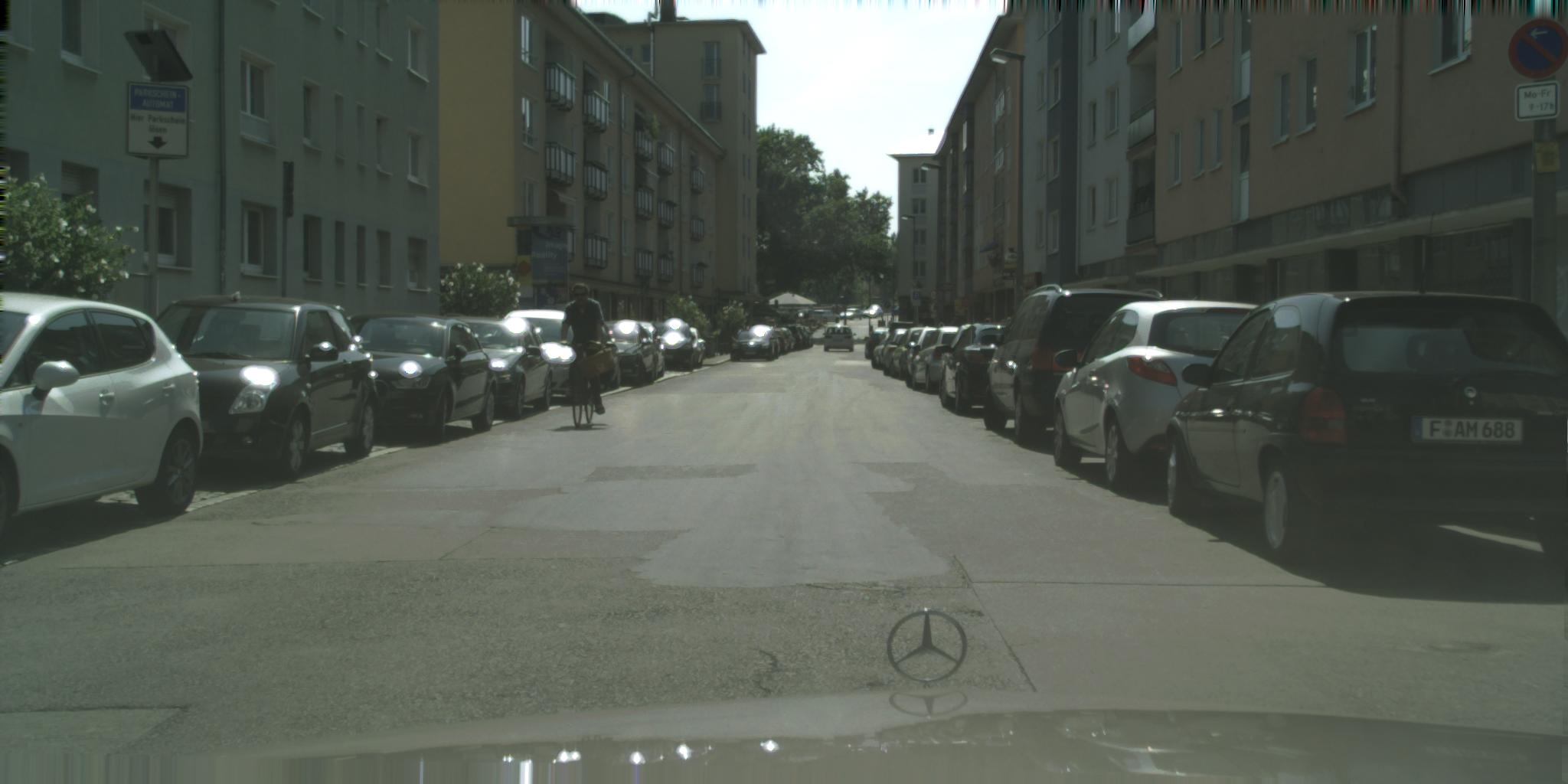}} &
        \setlength{\fboxsep}{0pt}  \fbox{\includegraphics[width=\lennine,keepaspectratio]{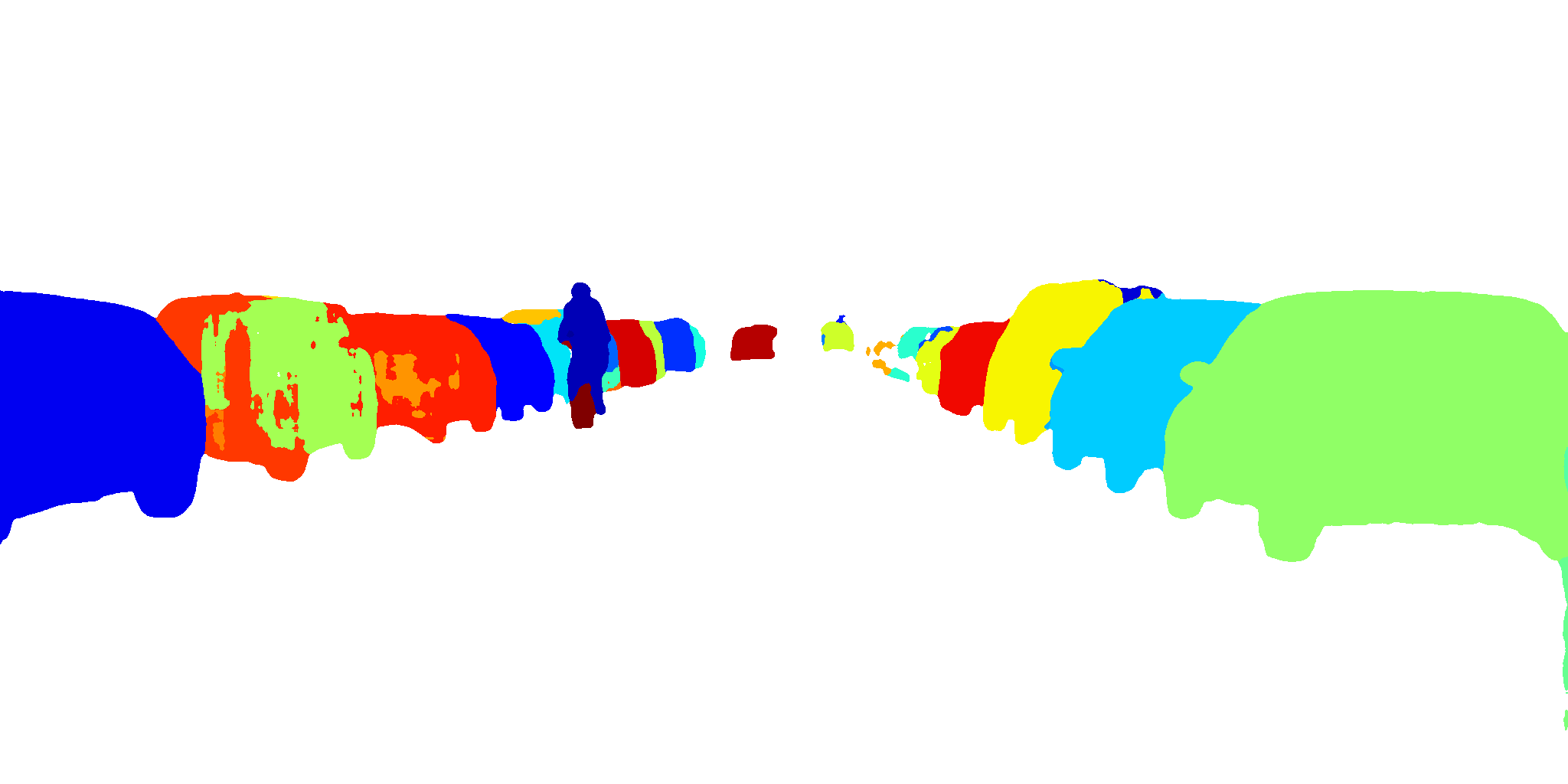}} & 
        \setlength{\fboxsep}{0pt}  \fbox{\includegraphics[width=\lennine,keepaspectratio]{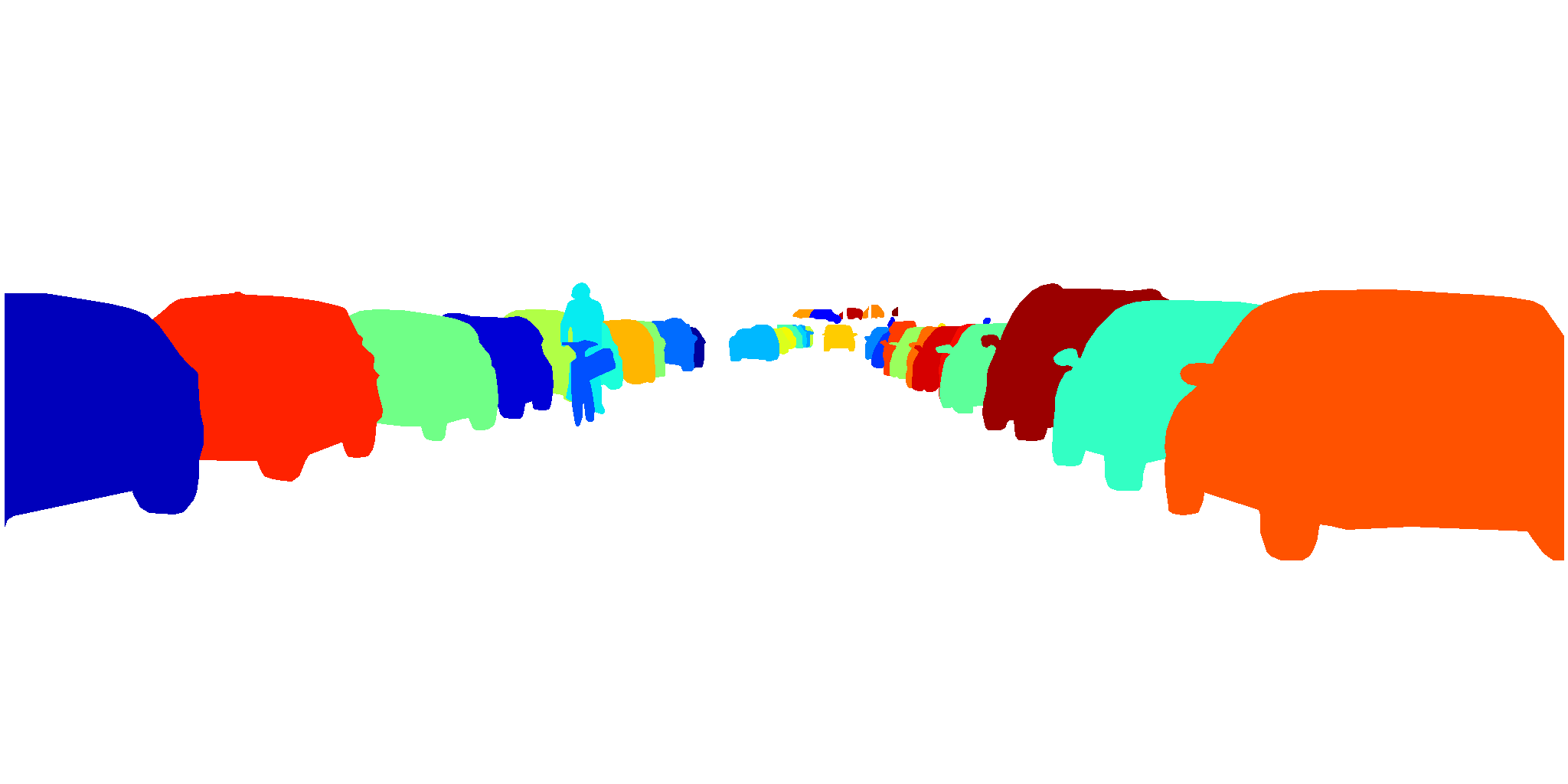}}
    \\
\end{tabular}
\end{center}
   \caption{Examples from the CityScapes validation set. In the middle column, we show the instances recovered by deep coloring. The right row contains ground truth instance segmentation. The bottom row show typical failure mode, where instances that were too small were not recovered. }
\label{fig:city}
\end{figure*}

\begin{table}

    \caption{Semantic instance segmentation using coloring. Semantic segmentation results (mean IoU) and semantic instance segmentation results (AP) on Cityscapes validation set are reported for classes with instances. The U-Net based architecture is used. See text for the discussion of the approaches. }
    \label{tab:cityscapes_semantic}
    \centering
    %\begin{tabular}{|c|cccccccc|c|}
    %\hline
    %Method & person & rider & car & truck & bus & train & motorcycle & bicycle & mean \\
    %\hline 
    %U-net + coloring & 63.2 & 49.9 & 87.8 & 43.7 & 60.7 & 43.2 & 34.6 & 55.4 & 54.8\\
    %U-net + coloring + semantic & 73.1 & 54.3 & 91.1 & 58.0 & 70.6 & 69.2 & 42.7 & 44.3 & 62.9\\
    %U-net + semantic  & 75.0 & 56.6 & 91.1 & 64.7 & 78.0 & 64.3 & 47.7 & 43.6 & 65.1\\
    %PSPNet~\cite{zhao2016pyramid} & 78.4 & 63.9 & 93.0 & 73.2 & 85.3 & 74.1 & 60.7 & 73.0 & 75.2\\  
    \begin{tabular}{|l|cc|}
    \hline
    Approach  &  mean IoU & AP \\
    \hline 
    Coloring with class-specific colors &  54.8 & 19.1\\
    + separate semantic head &  62.9 & 21.2\\  
    + fusion with PSP-Net & 75.2 & 29.7 \\
    \hline
    Semantic segmentation only (UNet)  & 65.1 & NA \\
    Semantic segmentation only (PSPNet)  & 75.2 & NA \\
    \hline
    \end{tabular}

\end{table}

\subsection{Cityscapes}
The Cityscapes dataset is focused on semantic understanding of urban street scenes for autonomous driving. This dataset has 2975 training images, 500 validation images and 1525 test images; for each image a ground truth semantic and instance segmentation are provided. We again used U-net like network architecture with several modifications. Firstly, we added batch normalization layers after each convolution layer. Since having large receptive field is crucial for us, we placed a PSP-module~\cite{zhao2016pyramid} in the bottleneck of the network. Finally, we added an extra block with two convolution layers and max-pooling at the beginning of the encoder and a corresponding block at the end of the decoder. The latter allowed us to increase the input image resolution while keeping the same network memory consumption. For data augmentation, we adopt random left-right mirror and random crop and resize with scale factor of up to 2. All training images were downsampled to have size 512x1024. We set margin $m$ to 40 pixels and the halo weight $\mu$ to 16. The minimal component size threshold $\tau$ was set to $40$ pixels, and the merging threshold $\rho$ was set to $20$ pixels.

\textbf{Combining semantic segmentation and instance segmentation.} The Cityscapes benchmark as well as many other tasks requires to perform instance segmentation and semantic segmentation at the same time. In principle, our approach allows to interleave semantic segmentation with instance segmentation directly by assigning different sets of colors to different classes (\textit{coloring with class specific colors}). We have assigned the following number of channels to each class: one channel to background, seven channels to `person', four channels to `rider', seven channels to `car', three channels to `truck', three channels to `bus', two channels to `train', three channels to `motorcycle', and six channels to `bicycle`. In total, 36 colors was used. The assignment was done heuristically and was not optimized in any way. At training time, the coloring process \eq{coloring} only considers colors allocated to the ground truth class. 
As the Cityscapes protocol requires assignment of instance confidence scores in order to compute the AP (average precision) measure, we have used the mean color probability $y[c,p]$ over all pixels in an instance as this score.

In the case of the Cityscapes benchmark, we have found that coloring with class specific colors utilized network capacity in a suboptimal way (\tab{cityscapes_semantic}) leading to poor performance both in terms of instance segmentation and semantic segmentation. Note that this observation is specific to Cityscapes benchmark, where the performance is bound by the capacity of the network, and may not translate to smaller problems with fewer semantic classes. While the accuracy of pixel-level semantic segmentation is of tangential interest to us, it is clear that a system with low per-pixel semantic segmentation accuracy cannot achieve high semantic instance segmentation accuracy. Hence, a system with higher per-pixel semantic segmentation is needed.

To improve the performance, we have added a separate ``head'' that performs semantic segmentation and is trained with standard pixel-level cross-entropy loss. At test time, we have performed the following fusion between the coloring head results and the pixel-level semantic segmentation head results. We took the connected components of different semantic classes as predicted by the semantic head. For each connected component, we assigned each pixel to its maximal color (among the colors assigned to this class) as predicted by the coloring head.
 This procedure breaks the components of the same class into instances. The confidence score of each instance for the AP computation is computed as mean product of the color assignment probabilities $y[c,p]$ and the semantic class probabilities over all pixels in the object. The resulting architecture (\textit{+ separate semantic head} in \tab{cityscapes_semantic}) worked much better in terms of the pixel-level semantic segmentation (mean IoU), which lead to the increase in the semantic instance segmentation score (AP).  We visualize the output of the coloring head in~\fig{city}. Furthermore, \fig{challenging} provides more detailed break-out of a single very challenging example. 

\setlength{\lennine}{80pt}

\begin{figure*}
\begin{center}
\setlength\tabcolsep{2pt}
\begin{tabular}{cccccc}
\includegraphics[width=\lennine,keepaspectratio]{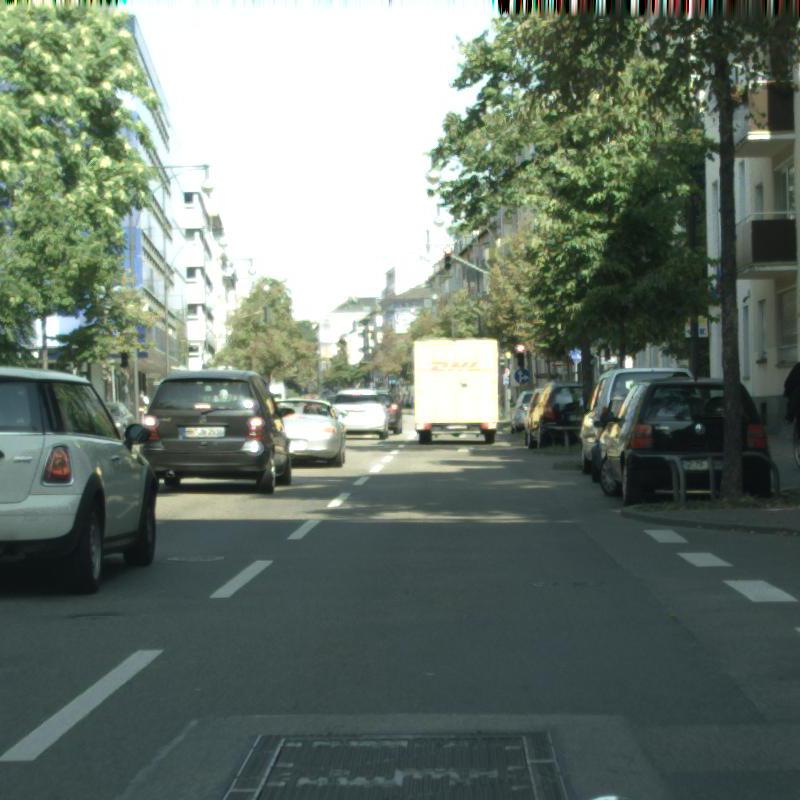} &
\includegraphics[width=\lennine,keepaspectratio]{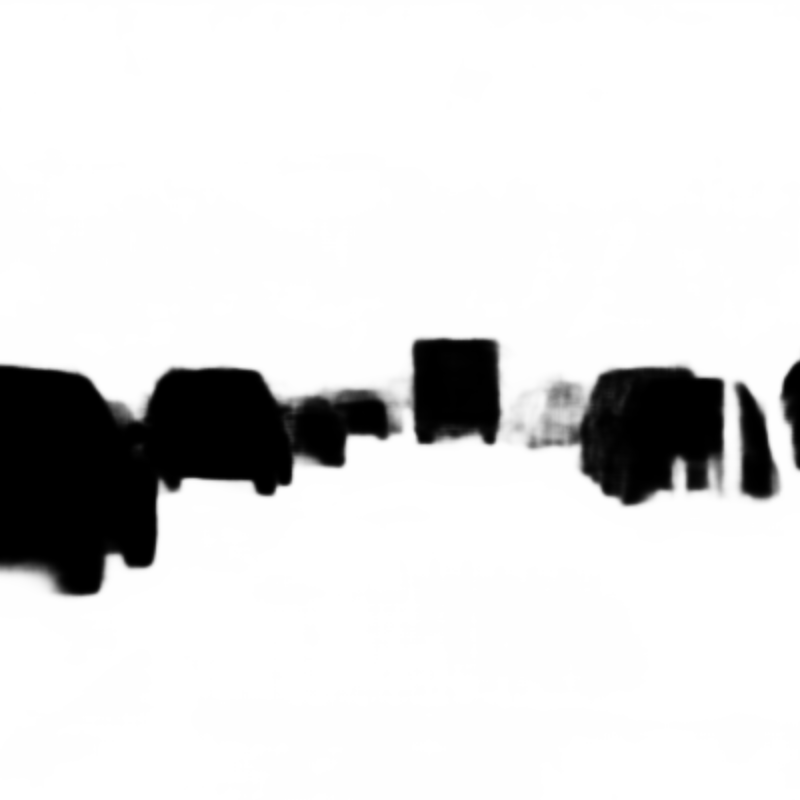} &
\includegraphics[width=\lennine,keepaspectratio]{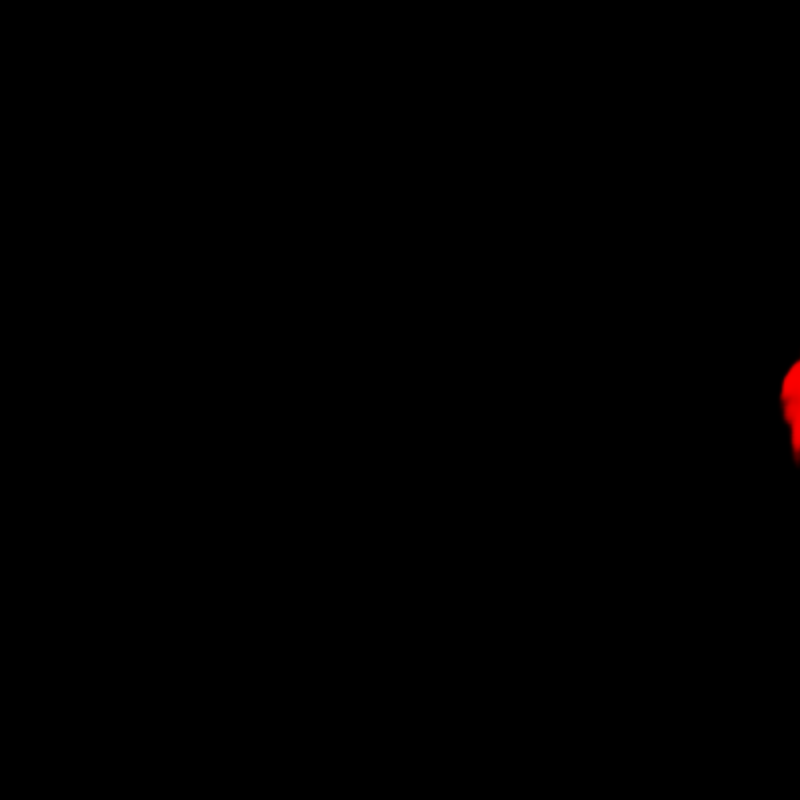} &
\includegraphics[width=\lennine,keepaspectratio]{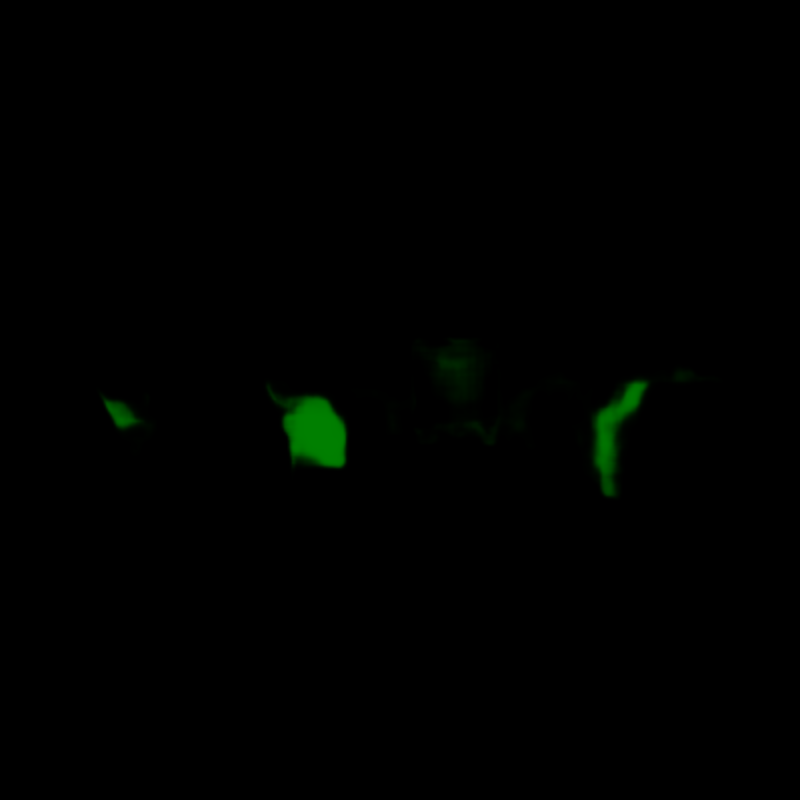} &
\includegraphics[width=\lennine,keepaspectratio]{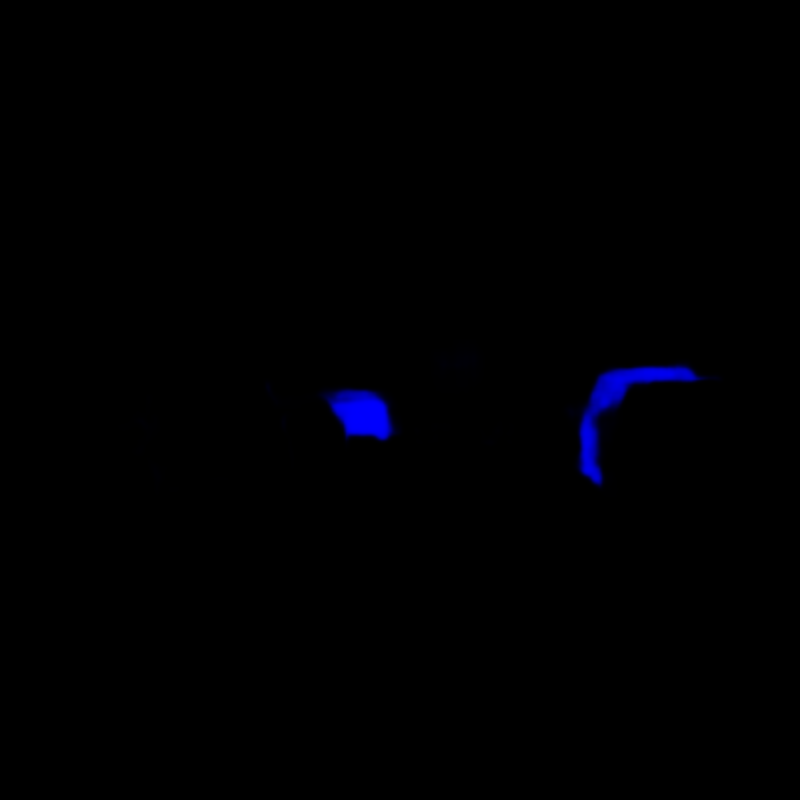} &
\includegraphics[width=\lennine,keepaspectratio]{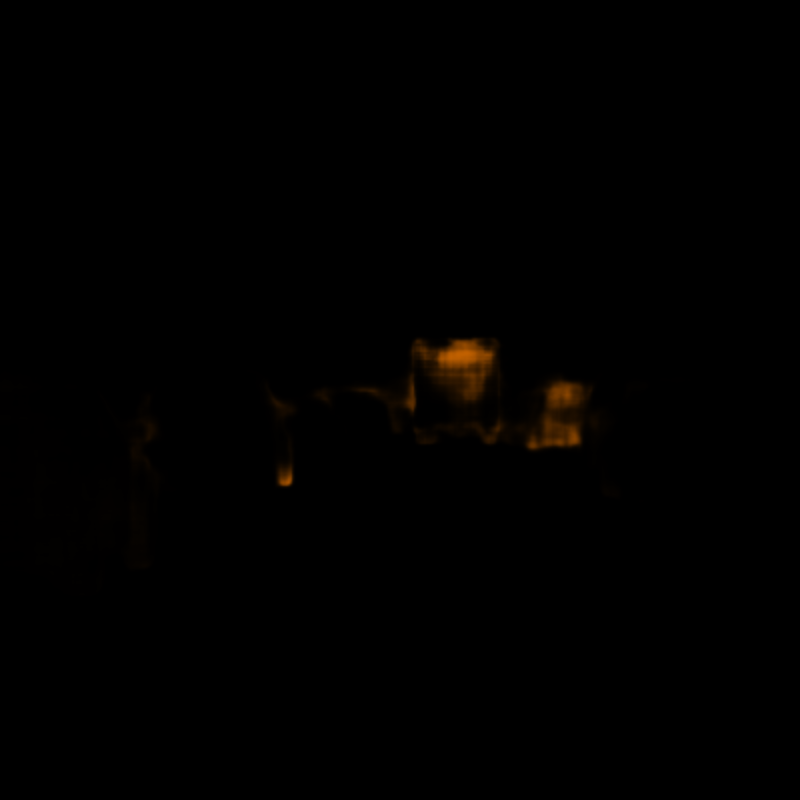} \\

\includegraphics[width=\lennine,keepaspectratio]{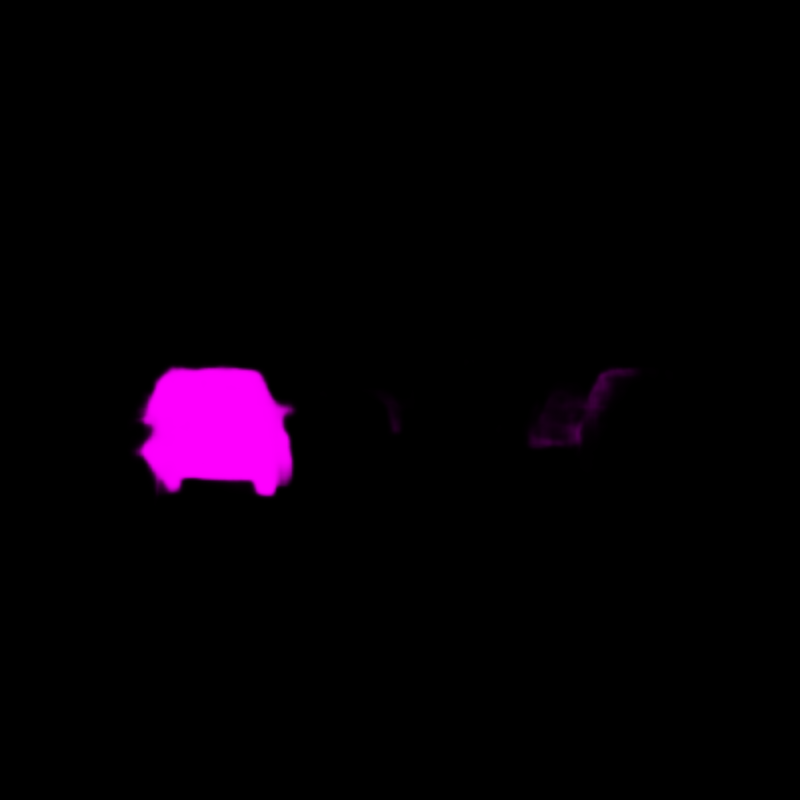} &
\includegraphics[width=\lennine,keepaspectratio]{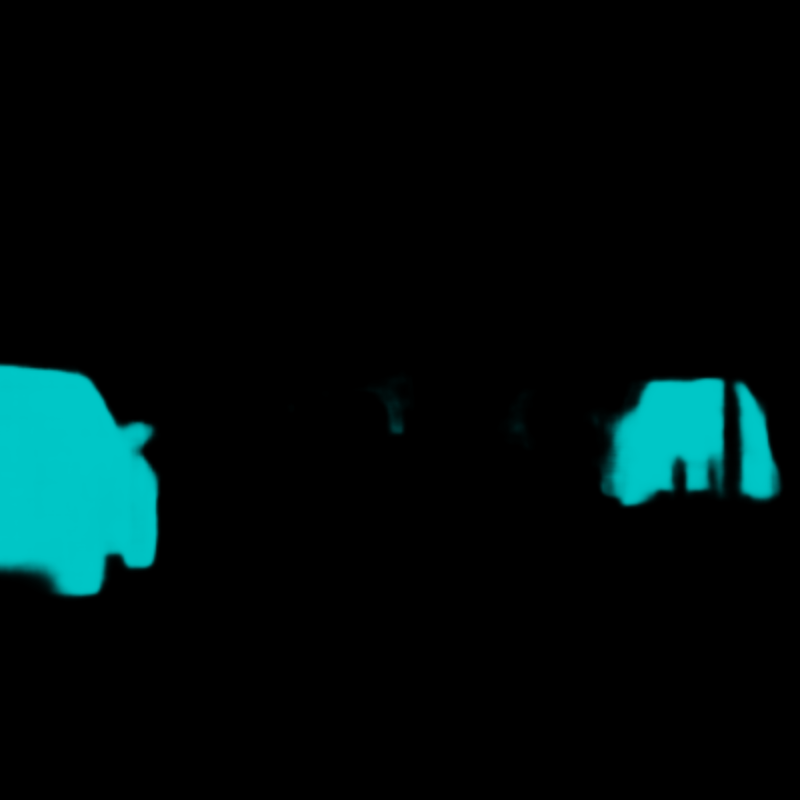} &
\includegraphics[width=\lennine,keepaspectratio]{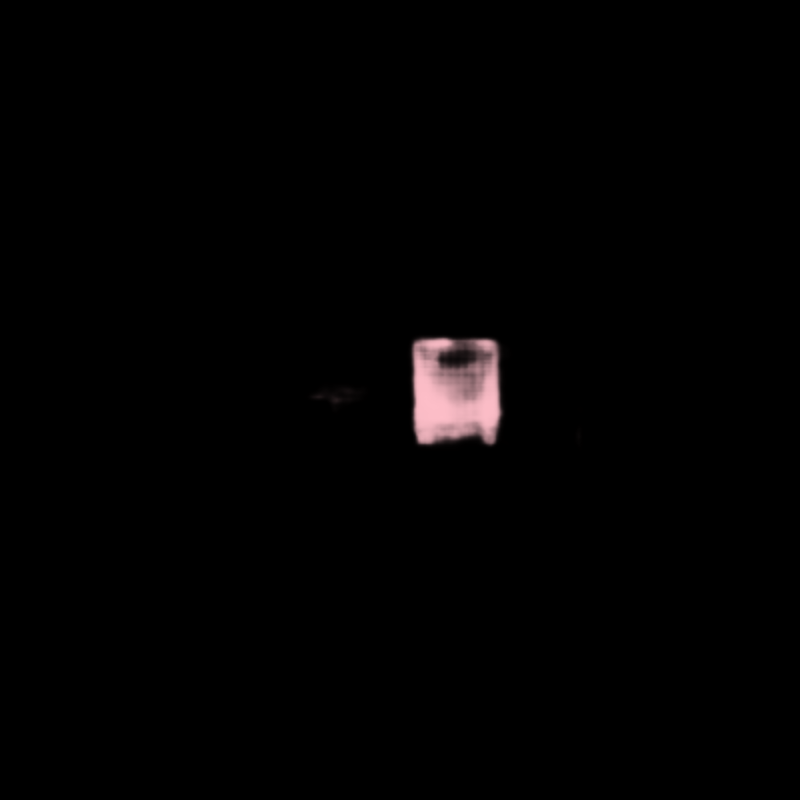} &
\includegraphics[width=\lennine,keepaspectratio]{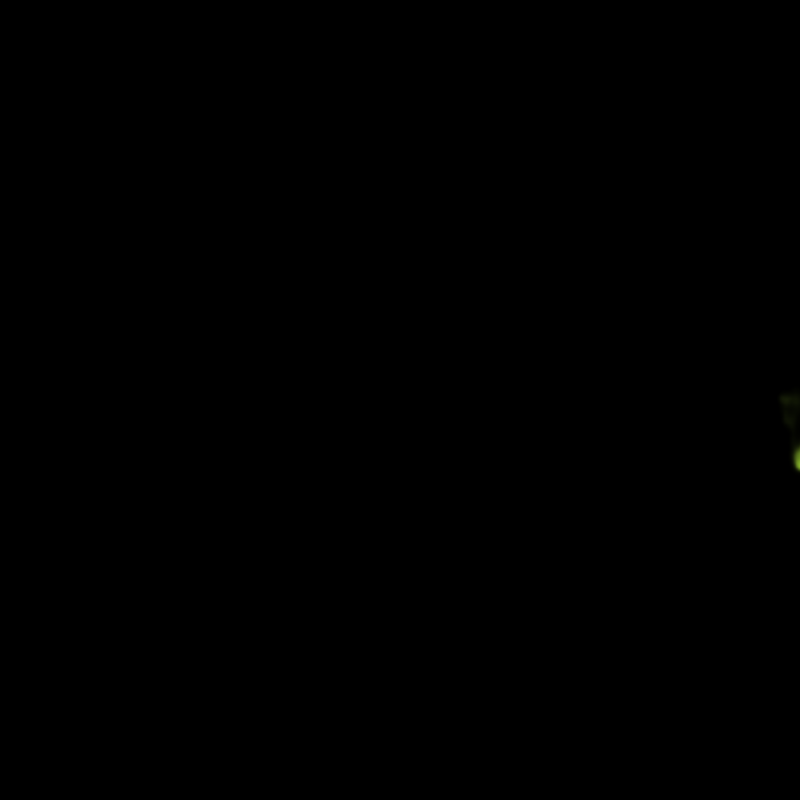} &
\includegraphics[width=\lennine,keepaspectratio]{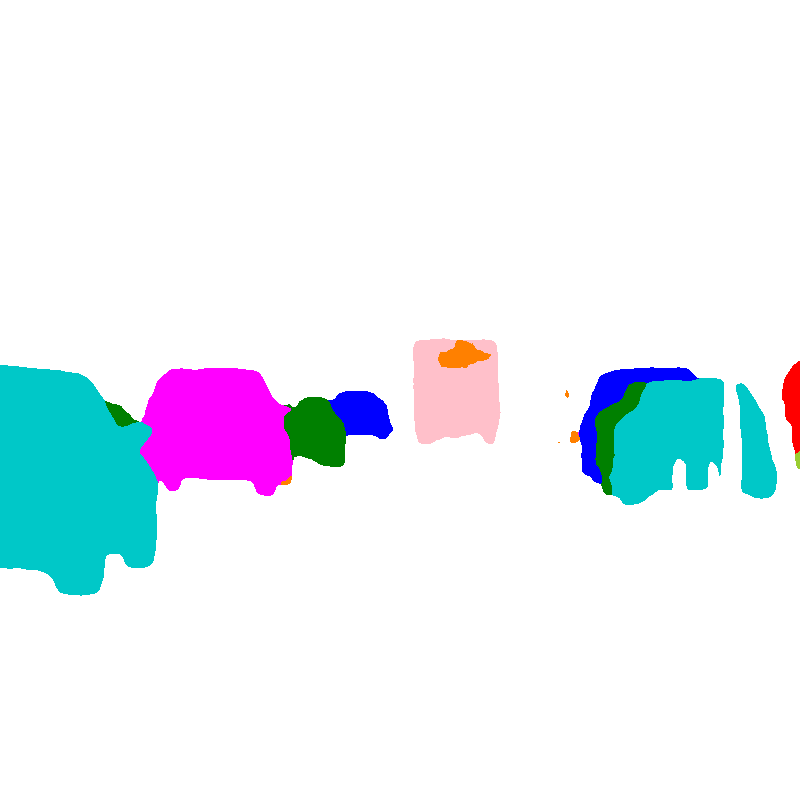} &
\includegraphics[width=\lennine,keepaspectratio]{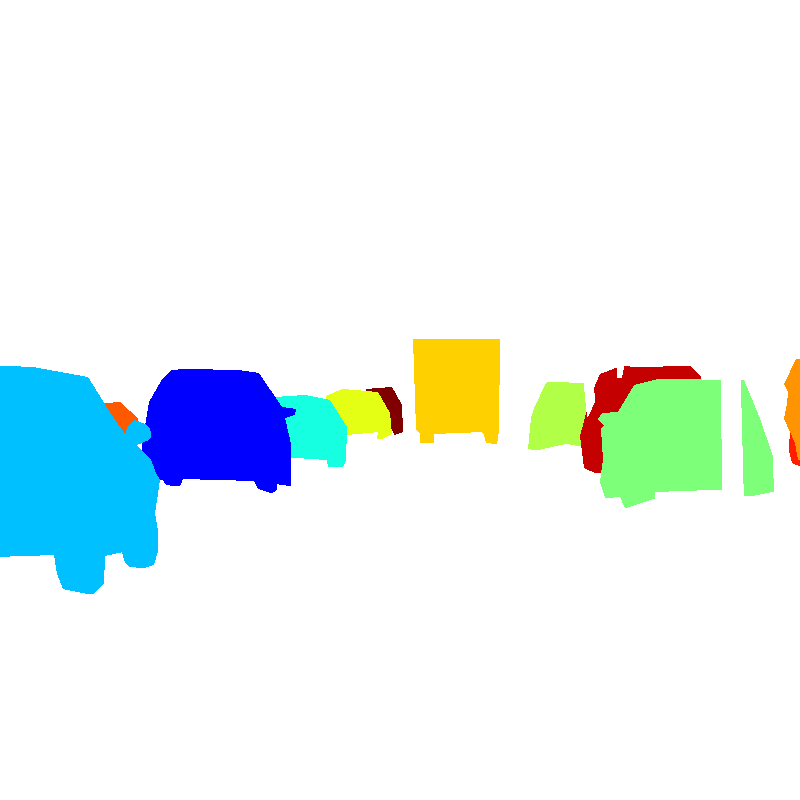}
\end{tabular}
\end{center}
   \caption{Processing a challenging scene fragment using our approach. The maps with strong activations are shown next to the input image. The bottom-left two images correspond to the map of obtained instances and the ground truth. A significant number of instances has been recovered despite crowding and small object sizes.}
\label{fig:challenging}
\end{figure*}

In terms of mean IoU, the two-head model performed slightly worse compared to the ``vanilla'' semantic segmentation model that uses semantic head only and puts all capacity on semantic segmentation (\textit{semantic segmentation} in \tab{cityscapes_semantic}). This is expected as in the case of the two head architecture some of the backbone capacity is allocated by the training process to the instance separation process.

\begin{table*}
    \centering
    \caption{Left -- results on the Cityscapes test set for methods using `fine'(only) train set. Standard metrics are reported (average precision AP, average precision more that 50\% of intersection over union and average precision within a certain distance). The performance of the proposed method (deep coloring) is competitive. Right -- the break-out of the performance of deep coloring over classes.  }
    \label{tab:cityscapes}    
   
    \begin{tabular}{cc}
        \begin{tabular}{|l|l l l l|}
        \hline
        &\small{AP}&\small{AP50\%}&\small{AP100m}&\small{AP50m}\\
        \hline
        %R-CNN+MCG\cite{cordts2016cityscapes} &4.6&12.9&7.7&10.3\\
        %JGD\cite{levinkov2017joint} &9.8&23.2&16.8&20.3\\
        Bound.-aware\cite{hayder2017boundary}&17.4&36.7&29.3&34.0\\
        Discriminative\cite{de2017semantic}&17.5&35.9&27.8&31.0\\
        DWT\cite{bai2016deep}&19.4&35.3&31.4&36.8\\
        Pixelwise DIN\cite{arnab2017pixelwise}&23.4&45.2&36.8&40.9\\
        SGN\cite{liu2017sgn}&25.0&44.9&38.9&44.5\\
        Mask R-CNN\cite{he2017mask}&26.2&49.9&37.6&40.1\\
        PolygonRNN++\cite{acuna2018efficient}&25.5&45.5&39.3&43.4\\
        PANet\cite{liu2018path}&\textbf{31.8}&\textbf{57.1}&\textbf{44.2}&\textbf{46.0}\\
        \hline
        \textbf{Deep coloring (+PSPNet semantic segm.)}&25.2&46.4&39.3&44.4 \\
        \hline
        \end{tabular}
        \quad&\quad
    \begin{tabular}{|c|cc|}
    \hline
    class & AP & AP50\%\\
    \hline 
    person &22.7  & 47.9\\
    rider & 21.6 &  49.5 \\
    car  &  40.8 &  63.1\\
    truck &  22.5  & 34.0 \\  
    bus & 33.6 &  46.5  \\ 
    train &  30.0  & 54.3 \\
    motorcycle & 18.0 &  42.8\\
    bicycle & 12.5 &  33.1  \\
    \hline
    \textbf{mean} &  25.2 & 46.4 \\
    \hline
    \end{tabular}
    \end{tabular}    

\end{table*}

\textbf{Maximizing performance.} The U-Net backbone is not competitive in terms of achievable semantic segmentation compared to more modern architectures such as PSPNet~\cite{zhao2016pyramid}, as the gap between the semantic segmentation performance using U-Net and PSP-Net is around $10\%$ mIoU. To achieve the optimal performance in terms of semantic instance segmentation, we have taken the two-head U-Net based architecture and evaluated the variant where the output of the coloring head is fused with the semantic segmentation obtained with the PSP-Net~\cite{zhao2016pyramid}. The training process for the architecture remains the same but at test time the output of the semantic head is discarded and replaced with the output of the ``off-the-shelf'' PSP-Net. This achieves a strong improvement on the validation set of the Cityscapes (\tab{cityscapes_semantic}).

The results of this variant on the Cityscapes test set are in \tab{cityscapes}. Our method performs favourably compared to the methods of the same group presented recently \cite{bai2016deep} or developed in parallel \cite{de2017semantic}.

\section{Conclusion}
We have developed a new approach for instance segmentation (\textit{deep coloring}). The approach reduces instance segmentation to the task of instance classification. The latter task can be accomplished using standard deep convolutional architectures for semantic segmentation. We have suggested a new simple rule for the dynamic coloring at training time, which coordinates the coloring process with the subsequent connected component extraction at test time. In a nutshell, the rule enforces all pixels of the same object to take the same color, while also enforcing pixels belonging to different but adjacent object  instances to take different colors. We have also shown that the proposed network can be trained end-to-end to perform two tasks simultaneously: background segmentation and instance segmentation.
The approach was shown to work well on two distinct datasets (CVPPP and CityScapes) and has been also tried on a dataset with the large number of instances (E.Coli dataset). PyTorch~\cite{pytorch} code is available at \texttt{https://github.com/kulikovv/DeepColoring} and a TensorFlow~\cite{tensorflow2015-whitepaper} implementation will be available upon publication.

% or


% \appendices

% % you can choose not to have a title for an appendix
% % if you want by leaving the argument blank
% \section{}

\ifCLASSOPTIONcompsoc
  \section*{Acknowledgments}
\else
  \section*{Acknowledgments}
\fi

This work was supported by the Skoltech NGP Program (Skoltech-MIT joint project).

\ifCLASSOPTIONcaptionsoff
  \newpage
\fi



%\bibliographystyle{IEEEtran}
\bibliographystyle{IEEEtran}
\bibliography{refs}

\begin{thebibliography}{10}
\providecommand{\url}[1]{#1}
\csname url@samestyle\endcsname
\providecommand{\newblock}{\relax}
\providecommand{\bibinfo}[2]{#2}
\providecommand{\BIBentrySTDinterwordspacing}{\spaceskip=0pt\relax}
\providecommand{\BIBentryALTinterwordstretchfactor}{4}
\providecommand{\BIBentryALTinterwordspacing}{\spaceskip=\fontdimen2\font plus
\BIBentryALTinterwordstretchfactor\fontdimen3\font minus
  \fontdimen4\font\relax}
\providecommand{\BIBforeignlanguage}[2]{{%
\expandafter\ifx\csname l@#1\endcsname\relax
\typeout{** WARNING: IEEEtran.bst: No hyphenation pattern has been}%
\typeout{** loaded for the language `#1'. Using the pattern for}%
\typeout{** the default language instead.}%
\else
\language=\csname l@#1\endcsname
\fi
#2}}
\providecommand{\BIBdecl}{\relax}
\BIBdecl

\bibitem{long2015fully}
J.~Long, E.~Shelhamer, and T.~Darrell, ``Fully convolutional networks for
  semantic segmentation,'' in \emph{Proceedings of the IEEE Conference on
  Computer Vision and Pattern Recognition}, 2015, pp. 3431--3440.

\bibitem{ronneberger2015u}
O.~Ronneberger, P.~Fischer, and T.~Brox, ``U-net: Convolutional networks for
  biomedical image segmentation,'' in \emph{International Conference on Medical
  Image Computing and Computer-Assisted Intervention}.\hskip 1em plus 0.5em
  minus 0.4em\relax Springer, 2015, pp. 234--241.

\bibitem{yu2015multi}
F.~Yu and V.~Koltun, ``Multi-scale context aggregation by dilated
  convolutions,'' \emph{arXiv preprint arXiv:1511.07122}, 2015.

\bibitem{rigamonti2013learning}
R.~Rigamonti, A.~Sironi, V.~Lepetit, and P.~Fua, ``Learning separable
  filters,'' in \emph{Proceedings of the IEEE Conference on Computer Vision and
  Pattern Recognition}, 2013, pp. 2754--2761.

\bibitem{chollet2016xception}
F.~Chollet, ``Xception: Deep learning with depthwise separable convolutions,''
  \emph{arXiv preprint arXiv:1610.02357}, 2016.

\bibitem{cordts2016cityscapes}
M.~Cordts, M.~Omran, S.~Ramos, T.~Rehfeld, M.~Enzweiler, R.~Benenson,
  U.~Franke, S.~Roth, and B.~Schiele, ``The cityscapes dataset for semantic
  urban scene understanding,'' in \emph{Proceedings of the IEEE Conference on
  Computer Vision and Pattern Recognition}, 2016, pp. 3213--3223.

\bibitem{lin2014microsoft}
T.-Y. Lin, M.~Maire, S.~Belongie, J.~Hays, P.~Perona, D.~Ramanan,
  P.~Doll{\'a}r, and C.~L. Zitnick, ``Microsoft coco: Common objects in
  context,'' in \emph{European conference on computer vision}.\hskip 1em plus
  0.5em minus 0.4em\relax Springer, 2014, pp. 740--755.

\bibitem{romera2016recurrent}
B.~Romera-Paredes and P.~H.~S. Torr, ``Recurrent instance segmentation,'' in
  \emph{European Conference on Computer Vision}.\hskip 1em plus 0.5em minus
  0.4em\relax Springer, 2016, pp. 312--329.

\bibitem{he2017mask}
K.~He, G.~Gkioxari, P.~Dollar, and R.~Girshick, ``Mask r-cnn,'' in \emph{The
  IEEE International Conference on Computer Vision (ICCV)}, Oct 2017.

\bibitem{he2016deep}
K.~He, X.~Zhang, S.~Ren, and J.~Sun, ``Deep residual learning for image
  recognition,'' in \emph{Proceedings of the IEEE conference on computer vision
  and pattern recognition}, 2016, pp. 770--778.

\bibitem{paszke2016enet}
A.~Paszke, A.~Chaurasia, S.~Kim, and E.~Culurciello, ``Enet: A deep neural
  network architecture for real-time semantic segmentation,'' \emph{arXiv
  preprint arXiv:1606.02147}, 2016.

\bibitem{zhao2016pyramid}
H.~Zhao, J.~Shi, X.~Qi, X.~Wang, and J.~Jia, ``Pyramid scene parsing network,''
  in \emph{The IEEE Conference on Computer Vision and Pattern Recognition
  (CVPR)}, July 2017.

\bibitem{dai2016instance}
J.~Dai, K.~He, Y.~Li, S.~Ren, and J.~Sun, ``Instance-sensitive fully
  convolutional networks,'' in \emph{European Conference on Computer
  Vision}.\hskip 1em plus 0.5em minus 0.4em\relax Springer, 2016, pp. 534--549.

\bibitem{li2016fully}
Y.~Li, H.~Qi, J.~Dai, X.~Ji, and Y.~Wei, ``Fully convolutional instance-aware
  semantic segmentation,'' in \emph{The IEEE Conference on Computer Vision and
  Pattern Recognition (CVPR)}, July 2017.

\bibitem{hariharan2014simultaneous}
B.~Hariharan, P.~Arbel{\'a}ez, R.~Girshick, and J.~Malik, ``Simultaneous
  detection and segmentation,'' in \emph{European Conference on Computer
  Vision}.\hskip 1em plus 0.5em minus 0.4em\relax Springer, 2014, pp. 297--312.

\bibitem{chen2015multi}
Y.-T. Chen, X.~Liu, and M.-H. Yang, ``Multi-instance object segmentation with
  occlusion handling,'' in \emph{Proceedings of the IEEE Conference on Computer
  Vision and Pattern Recognition}, 2015, pp. 3470--3478.

\bibitem{arbelaez2014multiscale}
P.~Arbel{\'a}ez, J.~Pont-Tuset, J.~T. Barron, F.~Marques, and J.~Malik,
  ``Multiscale combinatorial grouping,'' in \emph{Proceedings of the IEEE
  conference on computer vision and pattern recognition}, 2014, pp. 328--335.

\bibitem{Everingham10}
M.~Everingham, L.~Van~Gool, C.~K.~I. Williams, J.~Winn, and A.~Zisserman, ``The
  pascal visual object classes (voc) challenge,'' \emph{International Journal
  of Computer Vision}, vol.~88, no.~2, pp. 303--338, Jun. 2010.

\bibitem{hochreiter1997long}
S.~Hochreiter and J.~Schmidhuber, ``Long short-term memory,'' \emph{Neural
  computation}, vol.~9, no.~8, pp. 1735--1780, 1997.

\bibitem{ren2016end}
M.~Ren and R.~S. Zemel, ``End-to-end instance segmentation with recurrent
  attention,'' in \emph{The IEEE Conference on Computer Vision and Pattern
  Recognition (CVPR)}, July 2017.

\bibitem{bai2016deep}
M.~Bai and R.~Urtasun, ``Deep watershed transform for instance segmentation,''
  in \emph{The IEEE Conference on Computer Vision and Pattern Recognition
  (CVPR)}, July 2017.

\bibitem{uhrig2016pixel}
J.~Uhrig, M.~Cordts, U.~Franke, and T.~Brox, ``Pixel-level encoding and depth
  layering for instance-level semantic labeling,'' in \emph{German Conference
  on Pattern Recognition}.\hskip 1em plus 0.5em minus 0.4em\relax Springer,
  2016, pp. 14--25.

\bibitem{de2017semantic}
B.~De~Brabandere, D.~Neven, and L.~Van~Gool, ``Semantic instance segmentation
  with a discriminative loss function,'' \emph{arXiv preprint
  arXiv:1708.02551}, 2017.

\bibitem{Xu15}
J.~Xu, A.~G. Schwing, and R.~Urtasun, ``Learning to segment under various forms
  of weak supervision,'' in \emph{Proceedings of the IEEE Conference on
  Computer Vision and Pattern Recognition}.\hskip 1em plus 0.5em minus
  0.4em\relax IEEE, 2015, pp. 3781--3790.

\bibitem{Zhang15}
W.~Zhang, S.~Zeng, D.~Wang, and X.~Xue, ``Weakly supervised semantic
  segmentation for social images,'' in \emph{Proceedings of the IEEE Conference
  on Computer Vision and Pattern Recognition}, 2015, pp. 2718--2726.

\bibitem{Kolesnikov16}
A.~Kolesnikov and C.~H. Lampert, ``Seed, expand and constrain: Three principles
  for weakly-supervised image segmentation,'' in \emph{Computer Vision - {ECCV}
  2016 - 14th European Conference, Amsterdam, The Netherlands, October 11-14,
  2016, Proceedings, Part {IV}}, 2016, pp. 695--711.

\bibitem{scharr2014annotated}
H.~Scharr, M.~Minervini, A.~Fischbach, and S.~A. Tsaftaris, ``Annotated image
  datasets of rosette plants,'' in \emph{European Conference on Computer
  Vision. Z{\"u}rich, Suisse}, 2014, pp. 6--12.

\bibitem{minervini2016finely}
M.~Minervini, A.~Fischbach, H.~Scharr, and S.~A. Tsaftaris, ``Finely-grained
  annotated datasets for image-based plant phenotyping,'' \emph{Pattern
  recognition letters}, vol.~81, pp. 80--89, 2016.

\bibitem{scharr2016leaf}
H.~Scharr, M.~Minervini, A.~P. French, C.~Klukas, D.~M. Kramer, X.~Liu,
  I.~Luengo, J.-M. Pape, G.~Polder, D.~Vukadinovic \emph{et~al.}, ``Leaf
  segmentation in plant phenotyping: a collation study,'' \emph{Machine vision
  and applications}, vol.~27, no.~4, pp. 585--606, 2016.

\bibitem{pape20143}
J.-M. Pape and C.~Klukas, ``3-d histogram-based segmentation and leaf detection
  for rosette plants.'' in \emph{ECCV Workshops (4)}, 2014, pp. 61--74.

\bibitem{yin2014multi}
X.~Yin, X.~Liu, J.~Chen, and D.~M. Kramer, ``Multi-leaf tracking from
  fluorescence plant videos,'' in \emph{Image Processing (ICIP), 2014 IEEE
  International Conference on}.\hskip 1em plus 0.5em minus 0.4em\relax IEEE,
  2014, pp. 408--412.

\bibitem{giuffrida2016learning}
M.~V. Giuffrida, M.~Minervini, and S.~A. Tsaftaris, ``Learning to count leaves
  in rosette plants,'' 2016.

\bibitem{hayder2017boundary}
Z.~Hayder, X.~He, and M.~Salzmann, ``Boundary-aware instance segmentation,'' in
  \emph{Conference on Computer Vision and Pattern Recognition (CVPR)}, no.
  EPFL-CONF-227439, 2017.

\bibitem{arnab2017pixelwise}
A.~Arnab and P.~H.~S. Torr, ``Pixelwise instance segmentation with a
  dynamically instantiated network,'' in \emph{The IEEE Conference on Computer
  Vision and Pattern Recognition (CVPR)}, July 2017.

\bibitem{liu2017sgn}
S.~Liu, J.~Jia, S.~Fidler, and R.~Urtasun, ``Sgn: Sequential grouping networks
  for instance segmentation,'' in \emph{Proceedings of the IEEE Conference on
  Computer Vision and Pattern Recognition}, 2017, pp. 3496--3504.

\bibitem{acuna2018efficient}
D.~Acuna, H.~Ling, A.~Kar, and S.~Fidler, ``Efficient interactive annotation of
  segmentation datasets with polygon-rnn++,'' \emph{arXiv preprint
  arXiv:1803.09693}, 2018.

\bibitem{liu2018path}
S.~Liu, L.~Qi, H.~Qin, J.~Shi, and J.~Jia, ``Path aggregation network for
  instance segmentation,'' \emph{arXiv preprint arXiv:1803.01534}, 2018.

\bibitem{pytorch}
``{PyTorch} tensors and dynamic neural networks in python with strong gpu
  acceleration,'' \url{http://pytorch.org}, accessed: 2017-11-15.

\bibitem{tensorflow2015-whitepaper}
\BIBentryALTinterwordspacing
M.~Abadi, A.~Agarwal, P.~Barham, E.~Brevdo, Z.~Chen, C.~Citro, G.~S. Corrado,
  A.~Davis, J.~Dean, M.~Devin, S.~Ghemawat, I.~Goodfellow, A.~Harp, G.~Irving,
  M.~Isard, Y.~Jia, R.~Jozefowicz, L.~Kaiser, M.~Kudlur, J.~Levenberg,
  D.~Man\'{e}, R.~Monga, S.~Moore, D.~Murray, C.~Olah, M.~Schuster, J.~Shlens,
  B.~Steiner, I.~Sutskever, K.~Talwar, P.~Tucker, V.~Vanhoucke, V.~Vasudevan,
  F.~Vi\'{e}gas, O.~Vinyals, P.~Warden, M.~Wattenberg, M.~Wicke, Y.~Yu, and
  X.~Zheng, ``{TensorFlow}: Large-scale machine learning on heterogeneous
  systems,'' 2015, software available from tensorflow.org. [Online]. Available:
  \url{https://www.tensorflow.org/}
\BIBentrySTDinterwordspacing

\end{thebibliography}

% \input{refs.bib}

% % \bibitem{IEEEhowto:kopka}
% % H.~Kopka and P.~W. Daly, \emph{A Guide to \LaTeX}, 3rd~ed.\hskip 1em plus
% %   0.5em minus 0.4em\relax Harlow, England: Addison-Wesley, 1999.

% \end{thebibliography}


% \begin{IEEEbiography}{Victor Kulikov}
% \end{IEEEbiography}

% \begin{IEEEbiographynophoto}{Victor Yurchenko}

% % insert where needed to balance the two columns on the last page with
% %\newpage

% \begin{IEEEbiographynophoto}{Victor Lempitsky}


%\vfill


\end{document}